\RequirePackage{fix-cm}
\documentclass[twocolumn]{svjour3}          % twocolumn
\smartqed  % flush right qed marks, e.g. at end of proof
\usepackage{graphicx}
\usepackage[misc]{ifsym}
\usepackage[margin=0.1cm]{caption}
\usepackage{amssymb}
\usepackage{natbib}
\newcommand{\methodname}{{DeepProposals}}

\newcommand{\ie}{\textit{i}.\textit{e}. }
\newcommand{\eg}{\textit{e}.\textit{g}. }

\begin{document}

\title{\methodname: Hunting Objects and Actions by Cascading Deep Convolutional Layers%\thanks{Grants or other notes
}


\author{
Amir Ghodrati \and
Ali Diba \and
Marco Pedersoli \and
Tinne Tuytelaars \and
Luc Van Gool
}


\institute{A. Ghodrati (\Letter) \and A. Diba \and T. Tuytelaars \and L. Van Gool \at
              KU Leuven, ESAT-PSI, iMinds \\
              \email{amir.ghodrati@esat.kuleuven.be}            \\               
           \and
           M. Pedersoli \at
              LEAR project, Inria Grenoble Rhˆone-Alpes, LJK, CNRS, Univ. Grenoble Alpes, France
}

\date{Received: date / Accepted: date}

\maketitle

%------------------------------------------------------------------------
% Abstract
%------------------------------------------------------------------------
%%%%%%%%% ABSTRACT
\begin{abstract}
%In this paper we evaluate the quality of the activation layers of a convolutional neural network (CNN) for the generation of object proposals for object detection. 
%We start this paper with evaluating the quality of object detection proposals based on a sliding-window approach over different activation layers of a convolutional neural network (CNN). We notice that the final convolutional layers can find the object of interest with high recall but poor localization due to the coarseness of the feature maps. Instead, the first layers of the network can better localize the object of interest but with a reduced recall.
%Based on these observations we design 
In this paper, a new method for generating object and action proposals in images and videos is proposed. 
It builds on activations of different convolutional layers of a pretrained CNN,
combining the localization accuracy of the early layers with the high informativeness (and hence recall) of the later layers. 
To this end, we build an inverse cascade that, going backward from the later to the earlier convolutional layers of the CNN, selects the most promising locations and refines them in a coarse-to-fine manner. 
The method is efficient, because i) it re-uses the same features extracted for detection, ii) it aggregates features using integral images, and iii) it avoids a dense evaluation of the proposals thanks to the use of the inverse coarse-to-fine cascade. 
%We also extend the method to videos by linking object locations of consecutive frames so as to produce action proposals consistent over time.
The method is also accurate. We show that our \methodname~outperform most of the previously proposed object proposal and action proposal approaches and, when plugged into a CNN-based object detector, produce state-of-the-art detection performance.
\end{abstract}

%------------------------------------------------------------------------
% Introduction
%------------------------------------------------------------------------
\section{Introduction}
\label{sec:intro}

\begin{figure*}
%\centering
\begin{center}
%\scalebox{0.6}
{
\includegraphics[width=1\linewidth]{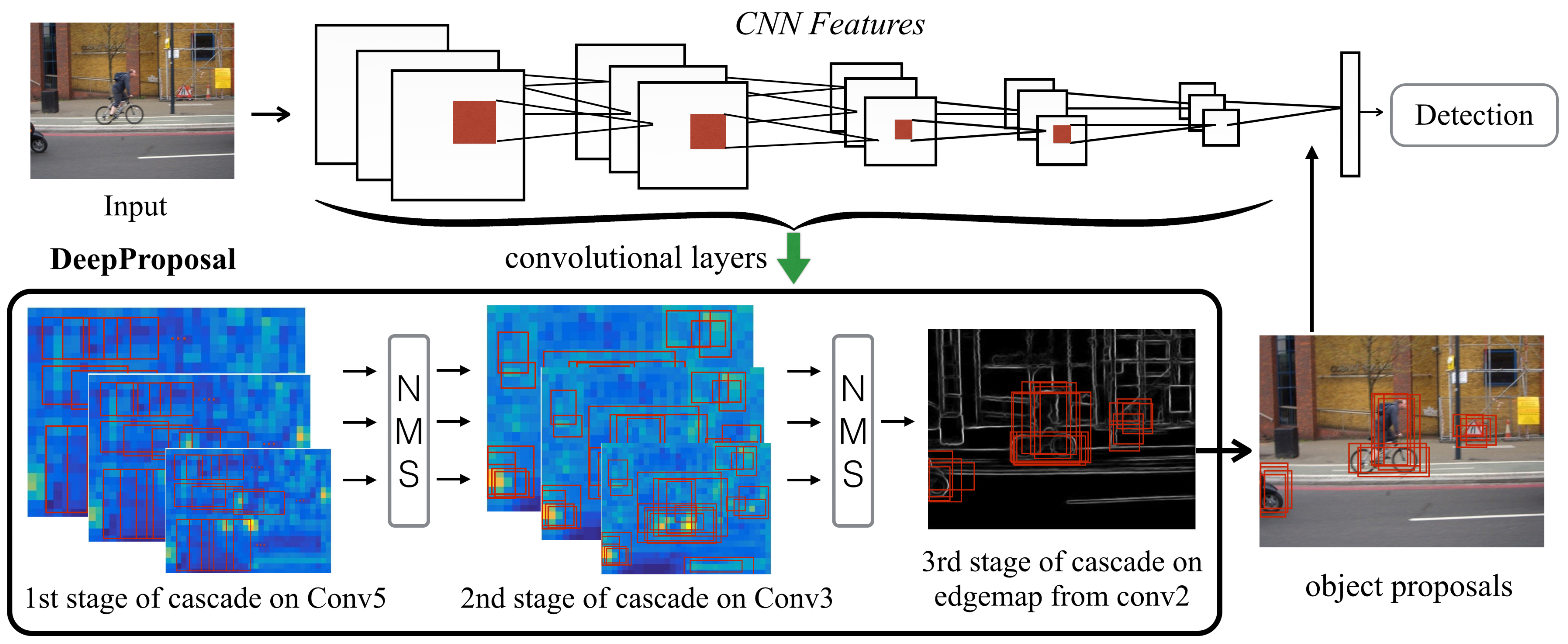}
}
\end{center}
\vspace{-0.4cm}
\caption{\methodname~pipeline. Our method uses the activation layers of a deep convolutional neural network in a coarse-to-fine inverse cascading to obtain proposals for object detection. %MP:we can speak about videos later or the frames of a video. 
Starting from a dense proposal sampling in the last convolutional layer (layer 5) we gradually filter out irrelevant boxes until reaching the initial layers of the net. In the last stage we use contours extracted from layer 2, to refine the proposals. Finally the generated boxes can be used within an object detection pipeline.}
\label{fig:overview}
\end{figure*}

%so far most proposals based on low level features
In recent years, the paradigm of generating a reduced set of window candidates to be evaluated with a powerful classifier has become very popular in object detection. Indeed, most of the recent state-of-the-art detection methods \citep{segfisherDet,ren2015faster,sppnet,regionlets} are based on such proposals. Furthermore, generating a limited number of proposals also helps weakly supervised object detection, 
%where learning can use few number of samples instead of strong supervision~\citett{weaklydetection,objlocalize}.
which consists of learning to localize objects without any bounding box annotations~\citep{objlocalize,weaklydetection}.

In fact, detection methods based on proposals can be seen as a two-stage cascade: first, the selection of a reduced set of promising and class-independent hypotheses, the proposals, and second, a class-specific classification of each hypothesis. Similarly to sliding window, this pipeline casts the detection problem to a classification problem. However, in contrast to sliding window, more powerful and time consuming detectors can be employed as the number of candidate windows is reduced. 

%high level features can also be interesting, especially if coming for free
Methods proposed in the literature for the generation of window candidates are based on two very different approaches. The first approach uses bottom-up cues like image segmentation \citep{MCG,selectivesearch}, object edges and contours \citep{edgebox} for window generation. The second approach is based on top-down cues which learn to separate correct object hypotheses from other possible window locations \citep{objectness, BING}. So far, the latter strategy seems to have inferior performance. In this paper we show that, with the proper features, accurate and fast top-down window proposals can be generated. %MP:I am not fully sure there are no methods based on top-down cues that performs well... reviewers can complain about this sentence.

%we evaluate CNN layers: high level=high recall poor localization, low level: good localization, worse recall
We consider for this task the convolutional neural network (CNN) ``feature maps'' extracted from the intermediate layers of Alexnet-like \citep{alexnet} networks. 
%and they seem general enough for giving state of the art results on several different tasks.
%We benefit from the fact that convolutional layers can deal with any input size because of their definition. So instead of re-sizing an image to pre-defined scale and then feed it to the CNN, we start from original image to have more suitable feature maps for generating object proposal. 
In the first part of this work we present a performance analysis of different CNN layers for generating proposals. More specifically, similarly to BING \citep{BING}, we select a reduced set of window sizes and aspect ratios and slide them on each possible location of the feature map generated by a certain CNN layer. The relevance (or objectness) of the windows is learned using a linear classifier. As the proposal generation procedure should be fast, we base the feature aggregation for each candidate window on average pooling, which can be computed in constant time using integral images \citep{violaintegral}. 

From this analysis we see that there is not a single best layer for candidate windows generation. Instead we notice that deeper layers, having a more semantic representation, perform very well in recalling the objects with a reduced set of hypotheses. Unfortunately, as noticed also for other tasks \citep{hyper}, they provide a poor localization of the object due to their coarseness.
In contrast, earlier layers are better in accurately localizing the object of interest, but their recall is reduced as they do not represent strong object cues.
%we propose a solution based on a cascade over layers
Thus, we conclude that, for a good window candidate generation, we should leverage multiple layers of the CNN. However, even with the very fast integral images for the feature extraction, evaluating all window locations at all feature layers is too expensive. Instead we propose a method based on a cascade starting from the last convolutional layer (layer 5) and going down with subsequent refinements until the initial layers of the net. As the flow of the cascade is inverse to the flow of the feature computation we call this approach an \textit{inverse cascade}. 
Also, as we start from a coarse spatial window resolution, and throughout the layers we select and spatially refine the window hypotheses until we obtain a reduced and spatially well localized set of hypotheses, it is a \textit{coarse-to-fine inverse cascade}. An overview of our approach, which we coined {\em DeepProposals}, is illustrated in Fig.~\ref{fig:overview}. 

%add a paragraph for extension to action proposals
In addition, we go beyond object proposals and extend the \methodname~framework (first proposed in \citet{ghodrati2015deepproposal}) to video, generating action proposals. To this end, we first apply the coarse-to-fine inverse cascade on each frame of a video. Then, we group the proposals into tubes, by imposing time continuity constraints, based on the assumption that the object of interest has a limited speed. We show that such proposals can provide excellent results for action localization.
% * <marcopede@gmail.com> 2016-04-12T21:06:55.472Z:
%
% This paragraph cuts the flow of the intro but I could not find a better place where to place it... AG:no any idea
%
% ^ <marcopede@gmail.com> 2016-04-12T21:07:33.903Z.

We evaluate the performance of the \methodname~in terms of recall vs. number of proposals as well as in terms of recall vs. object overlap. We show that in both evaluations the method is better than the current state of the art, and computationally very efficient.
However, the biggest gains are achieved when using the method as part of a CNN-based detector like one proposed in \citet{girshick15fastrcnn}. In this case the approach does not need to compute any feature, because it reuses the same features already computed by the CNN network for detection. Thus, we can execute the full detection pipeline at a very low computational cost.

This paper is an extension of our earlier work \citep{ghodrati2015deepproposal}. In this version we have included some additional related work and have compared it against some recent approaches. In addition, we have extended \methodname~for generating actions proposals in videos and have compared it with other state-of-the-art methods on two action datasets.

In the next section, we describe the related work.
Next, in section~\ref{sec:basic}, we analyze the quality of different CNN layers for window proposal generation. Section~\ref{sec:hierarchy} describes our inverse coarse-to-fine cascade and in section~\ref{sec:time_cont} its extension to action proposals. In section~\ref{sec:comparison} we compare quantitatively and qualitatively our method with the state-of-the-art, for both object and action proposal generation. Section~\ref{sec:conclusion} concludes the paper.

%------------------------------------------------------------------------
% Related Works
%------------------------------------------------------------------------
\section{Related work}
\label{sec:related}

\paragraph{\textbf{Object proposal methods}}

Object proposal generators aim at obtaining an accurate object localization with few object window hypotheses. These proposals can help object detection in two ways: searching objects in fewer locations to reduce the detector running time and/or using more sophisticated and expensive models to achieve better performance.

%Object proposal methods can be grouped mainly in three approaches. 
A first set of approaches measures the objectness of densely sampled windows (\ie how likely is it for an image window to represent an object)~\citep{objectness,BING,edgebox}.
\citet{objectness} propose a measure based on image saliency and other cues like color and edges to discriminate object windows from background. 
BING~\citep{BING} is a very fast proposal generator, obtained by training a classifier on edge features, but it suffers from low localization accuracy.
Moreover, ~\citet{crackBING} has shown that the BING classifier has minimal impact on locating objects and without looking at the actual image a similar performance can be obtained. %MP: I am a bit worried that this can be connected with the fact that in the intro we say that our window generation is similar to bing. We should explain here why in our case it works well!
Edgeboxes~\citep{edgebox} uses structural edges of \citet{structurededge}, a state-of-the-art contour detector, to compute proposal scores in a sliding window fashion without any parameter learning. For a better localization it uses a final window refinement step. %AG:cite possibly more methods (no need to explain them). %MP: mention somewhere also "Boosting Convolutional Features for Robust Object Proposals", it uses CNN features layer 1
Like these methods, our approach densely samples hypotheses in a sliding window fashion. However, in contrast to them, we use a hierarchy of high-to-low level features extracted from a deep CNN which has proven to be effective for object detection \citep{RCNN,regionlets}. 

An alternative approach to sliding-window methods are the segmentation-based algorithms. This approach extracts from the image multiple levels of bottom-up segmentation and then merges the generated segments in order to generate object proposals~\citep{MCG,CPMC,randomprime,selectivesearch}.
The first and most widely used segmentation-based algorithm is selective search~\citep{selectivesearch}. It hierarchically aggregates multiple segmentations in a bottom-up greedy manner without involving any learning procedure, but based on low level cues, such as color and texture. 
Multiscale Combinatorial Grouping (MCG)~\citep{MCG} extracts multiscale segmentations and merges them by using the edge strength in order to generate object hypotheses.
\citet{CPMC} propose to segment the object of interest based on graph-cut. It produces segments from randomly generated seeds. As in selective search, each segment represents a proposal bounding box. %AG: explain more clear what CPMC does
Randomized Prim's \citep{randomprime} uses the same segmentation strategy as selective search. However, instead of merging the segments in a greedy manner it learns the probabilities for merging, and uses those to speed up the procedure. Geodesic object proposals \citep{Geodesic} are based on classifiers that place seeds for a geodesic distance transform on an over-segmented image.
%Object proposals are defined by level sets of each of the distance transform from foreground/background segmentations \citet{Rodrigo14}.

Recently, following the great success of CNN in different computer vision tasks, CNN-based methods have been used to either generate proposals or directly regress the coordinates of the object bounding box. 
MultiBox~\citep{erhan2014scalable} proposes a network which directly regresses the coordinates of all object bounding boxes (without a sliding fashion approach) and assigns a confidence score for each of them in the image. However, MultiBox is not translation invariant and it does not share features between the proposal and detection networks i.e. it dedicates a network just for generating proposals. 

DeepMask~\citep{pinheiro2015learning} learns segmentation proposals by training a network to predict a class-agnostic mask for each image patch and an associated score. Same as MultiBox, they do not share features between the proposal generation and detection. Moreover, they need segmentation annotations to train their network.
OverFeat~\citep{sermanet2013overfeat} is a method which does proposal generation and detection in one-stage. In OverFeat, region-wise  features are extracted from a sliding window and are used to simultaneously determine the location and category of the objects. In contrast to it, our goal is to predict class-agnostic proposals which can be used in a second stage for class-specific detections.

Probably, the most similar to our work is the concurrent work of Region Proposal Network (RPN) proposed in \citet{ren2015faster}. RPN is a convolutional network that simultaneously predicts object bounds and objectness scores at each position. To generate region proposals, they slide a small network over the convolutional feature map. At each sliding-window location, they define $k$ reference boxes (anchors) at different scales and aspect ratios and predict multiple region proposals parameterized relative to the anchors. Similarly to us, RPN builds on the convolutional features of the detection network. However, we leverage low-level features in early layers of the network to improve the localization quality of proposals. In addition, in contrast to them, our method can build on any pre-trained network without the need to re-train it explicitly for the proposal generation task.

\paragraph{\textbf{Action proposal methods}} Action proposals are 3D boxes or temporal tubes extracted from videos that can be used for action localization, i.e. predicting the action label in a video and spatially localizing it. Also in this case, the main advantage of using proposals is to reduce the computational cost of the task and therefore make the method faster or allow for the use of more powerful classification approaches.
%The action proposal methods can be mainly divided to two groups.
The action proposal methods proposed in the literature to date mainly extend ideas originally developed for 2D object proposals in static images to 3D space. \citet{selectivesearch_v} is an extension of selective search~\citep{selectivesearch} to video. It extract super-voxels instead of super-pixels from a video and by hierarchical grouping it produces spatio-temporal tubes.

\citet{objectness_v} is an action proposal method inspired by the objectness method~\citep{objectness}, while a spatio-temporal variant of randomized Prime~\citep{randomprime} is proposed in~\citet{randprime_v}. Since most of those methods are based on a super-pixel segmentation approach as a pre-processing step, they are computationally very expensive. To avoid such computationally demanding pre-processing,~\citet{van2015apt} proposed Action Localization Proposals (APT) which use the same features used in detection to generate action proposals. %MP: we do also that right? What's the difference then?

Several action localization methods use 2D proposals in each frame but without generating intermediate action proposals at video-level.
Typically, they leverage 2D object proposals that are generated separately for each frame in order to find the most probable path of bounding boxes across time for each action class separately~\citep{gkioxari2015finding,tran2014video,weinzaepfel2015learning,yu2015fast}. Our method is similar to these works in spirit. However, these methods use class-specific detectors for action localization while we propose a class-agnostic method to generate a reduced set of action proposals. The idea of using class-agnostic proposals allows us to filter out many negative tubes with a reduced computational time which enables the use of more powerful classifiers in the final stage.

%------------------------------------------------------------------------
% Basic Approach
%------------------------------------------------------------------------
\section{CNN layers for proposals generation}
\label{sec:basic}
In this section we analyze the quality of the different layers of a CNN as features for window proposal generation. The window proposals ideally should cover all objects in an image. % and all actions in a static frame. 
% * <marcopede@gmail.com> 2016-04-12T21:43:08.410Z:
%
% I would not speak about actions at this point as the evaluation is done only on VOC07, which is not for action. AG:OK, makes sense
%
% ^.
To evaluate the baselines in this section, we use the PASCAL VOC 2007 dataset~\citep{pascal}.

\subsection{Basic Approach}
\label{subsec:basic}
%Given a CNN layer, we first explain the basic procedure for the candidate windows generation.
\paragraph{\textbf{Sliding window}}
Computing all possible boxes in a feature map of size $N \times N$ is in the order of $O(N^4)$ and therefore computationally unfeasible. Hence, similarly to \citet{BING}, we select a set of window sizes that best cover the training data in terms of size and aspect ratio and use them in a sliding window fashion over the selected CNN layer. 
This approach is much faster than evaluating all possible windows and avoids to select windows with sizes or aspect ratios different from the training data and therefore probably false positives.

For the selection of the window sizes, we start with a pool of windows $W_{all}$ in different sizes and aspect ratios $W_{all}:\{\omega|\omega \in \mathbb{Z}^2, \mathbb{Z}=[1..20]\}$. 
It is important to select a set of window sizes that gives high recall and at the same time produces well localized proposals. 
To this end, for each window size, we compute its recall with different Intersection over Union (\texttt{IoU}) thresholds and greedily pick one window size at a time that maximizes $\sum_{\alpha} recall(\texttt{IoU}>\alpha)$ over all the objects in the training set. Using this procedure, $50$ window sizes are selected for the sliding window procedure. In Fig.~\ref{fig:layer}~({\bf middle}) we show the maximum recall that can be obtained with the selected window sizes, which is an upper bound of the achievable recall of our method.

\paragraph{\textbf{Multiple scales}}
Even though it is possible to cover all possible objects using a sliding window on a single scale of feature map, it is inefficient since by using a single scale the stride is fixed and defined by the feature map resolution. For an efficient sliding window, the window stride should be proportional to the window size.
Therefore, in all the experiments we evaluate our set of windows on multiple scales. For each scale, we resize the image such that $min(w,h)=s$ where $s\in \{227, 300, 400, 600\}$. Note that the first scale is the network original input size.
%MP: In the multiple scale setting is not mentioned how you select the window sizes!

\paragraph{\textbf{Pooling}}
As the approach should be very fast we represent a window by the average pooling of the convolutional features that are inside the window. As averaging is a linear operation, after computing the integral image, the features of any proposal window can be extracted in a constant time.
Let $f(x,y)$ be the specific channel of the feature map from a certain CNN layer and $F(x,y)$ its integral image.
Then, average pooling $avg$ of a box defined by the top left corner $a=(a_x,a_y)$ and the bottom right corner $b=(b_x,b_y)$ is obtained as:

\small
\begin{equation}
	Avg(a,b)=\frac{F(b_x,b_y)-F(a_x,b_y)-F(b_x,a_y)+F(a_x,a_y)}{(b_x-a_x)(b_y-a_y)}.
\label{equ:int}    
\end{equation}
\normalsize

Thus, after computing the integral image, the average pooling of any box is obtained in a constant time that corresponds to summing $4$ integral values and dividing by the area of the box.

\iffalse
We also evaluate a max pooling strategy. Again, as one of the most important constraints in our approach is the speed, instead of using a real max pooling whose computational cost is linear in the number of elements present in a window, we approximate the max with a soft-max. The soft-max is defined as:
\begin{equation}
	softmax(a,b)= \frac{1}{\beta}\log \sum_{x=[a_x,b_x],y=[a_y,b_y]} e^{\beta f(x,y,c)}.
\end{equation}
The softmax, being a sum of exponentials can be computed with the integral image. First the features are exponentiated and then their integral image $E(x,y,c)$ is computed. Thus, when we want to compute the softmax of a certain window defined by $(a,b)$, we can use the same computation as in Eq.~\ref{equ:int}, but substituting $F(a,b)$ by its exponentiated version $E(a,b)$. Finally, we take the logarithm of the obtained feature and divide it by $\beta$. We have tested that with $\beta>100$ there is no difference between the softmax and the real-max. %For values lower than $1$ the function is in between average and max pooling.
\fi
%We can also extend this basic approach to include a pyramid representation that, even though slower, introduces more geometrical information in the description of a window. To this end, we compute the features with pyramid $1\times1$ and $2\times2$.

\paragraph{\textbf{Pyramid}}
One of the main cues used to detect general objects is the object boundaries. Using an approach based on average pooling can dilute the importance of the object boundaries because it discards any geometrical information among features.
Therefore, to introduce more geometry to the description of a window we consider a spatial pyramid representation~\citep{Spatialpyramid}.
It consists of dividing the proposal window into a number of same size sub-windows (\eg $2\times2$), and for each one build a different representation.

\paragraph{\textbf{Bias on size and aspect ratio}}
Objects tend to appear at specific sizes and aspect ratios. Therefore we add in the feature representation 3 additional dimensions $(w,h,w\times h)$ where $w$ and $h$ are the width and height of a window. This can be considered as an explicit kernel which lets the SVM learn which object sizes can be covered in a specific scale.
For the final descriptor, we normalize the pooled features and size-related features separately with $l_2$ norm.
%\paragraph{Normalization}

\paragraph{\textbf{Classifier}}
We train a linear classifier for each scale separately. For a specific scale, we randomly select at most 10 regions per object that overlap the annotation bounding boxes more than 70\%, as positive training data and 50 regions per image that overlap less than 30\% with ground-truth objects as negative data. In all experiments we use a linear SVM~\citep{LIBLINEAR} because of its simplicity and fast training. We did not test non-linear classifiers since they would be too slow for our approach.
%Details about the training of the classifier are given in section \ref{}.
%apply nms

\paragraph{\textbf{Non-maximum suppression}}
The ranked window proposals in each scale are finally reduced through a non-maximum suppression step. A window is removed if its \texttt{IoU} with a higher scored window is more than a threshold $\alpha$, which defines the trade-off between recall and accurate localization. So, this threshold is directly related to the \texttt{IoU} criteria that is used for evaluation (see sec \ref{subsec:eval_basic}). By tuning $\alpha$, it is possible to maximize recall at arbitrary \texttt{IoU} of $\beta$. Particularly, in this work we define two variants of our \methodname,~namely \methodname50 and \methodname70 for maximizing recall at \texttt{IoU} of $\beta=0.5$ and $\beta=0.7$ respectively by fixing $\alpha$ to $\beta+0.05$ (like~\citet{edgebox}). In addition, to aggregate boxes from different scales, we use another non-maximum suppression, fixing $\alpha=\beta$.

\subsection{Evaluation for object proposals}
\label{subsec:eval_basic}
\begin{figure*}
\begin{center}
%\scalebox{0.8}
%{
\begin{tabular}{ccc}
%\begin{center}
\includegraphics[width=0.32\linewidth]{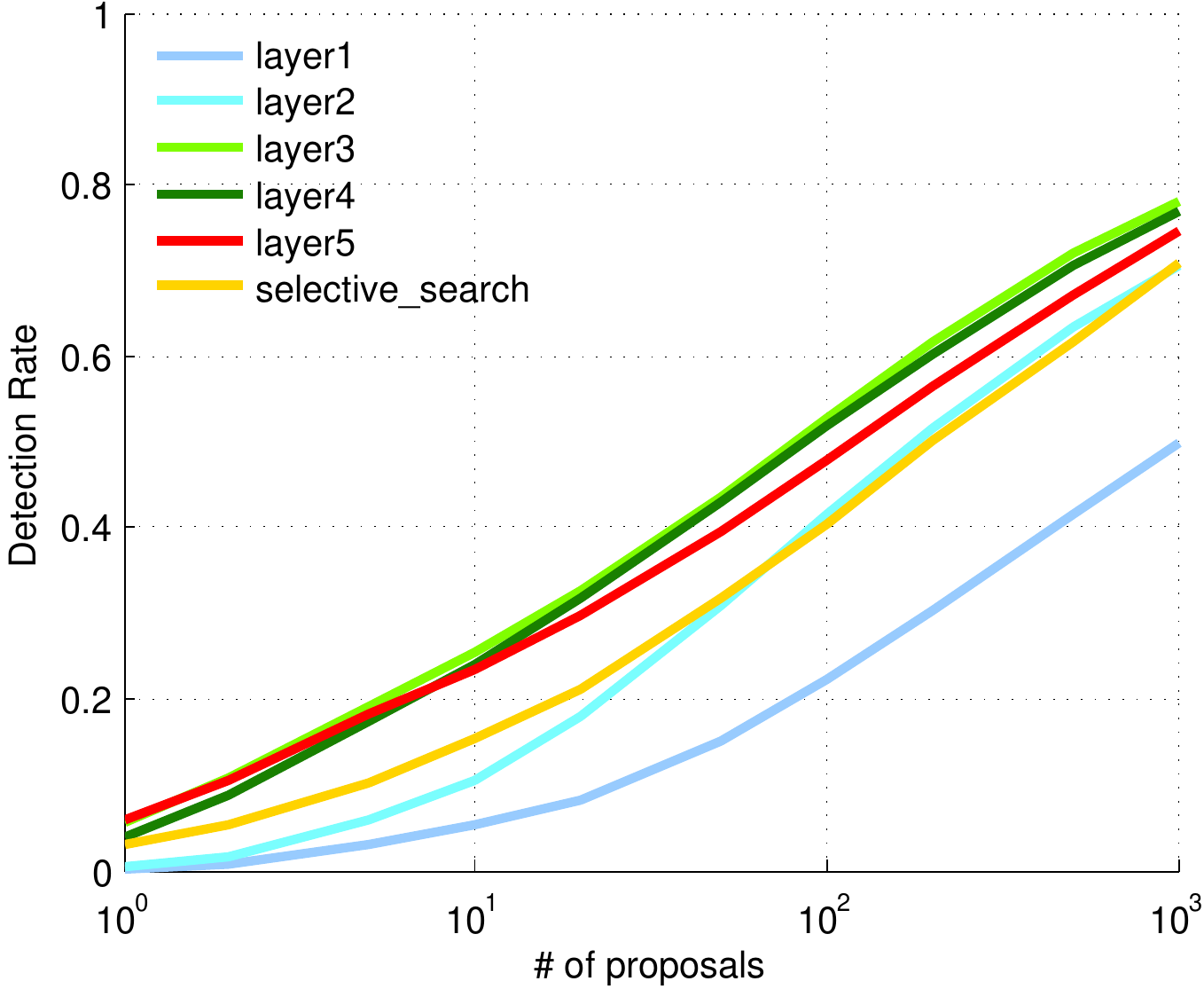}
&
\includegraphics[width=0.32\linewidth]{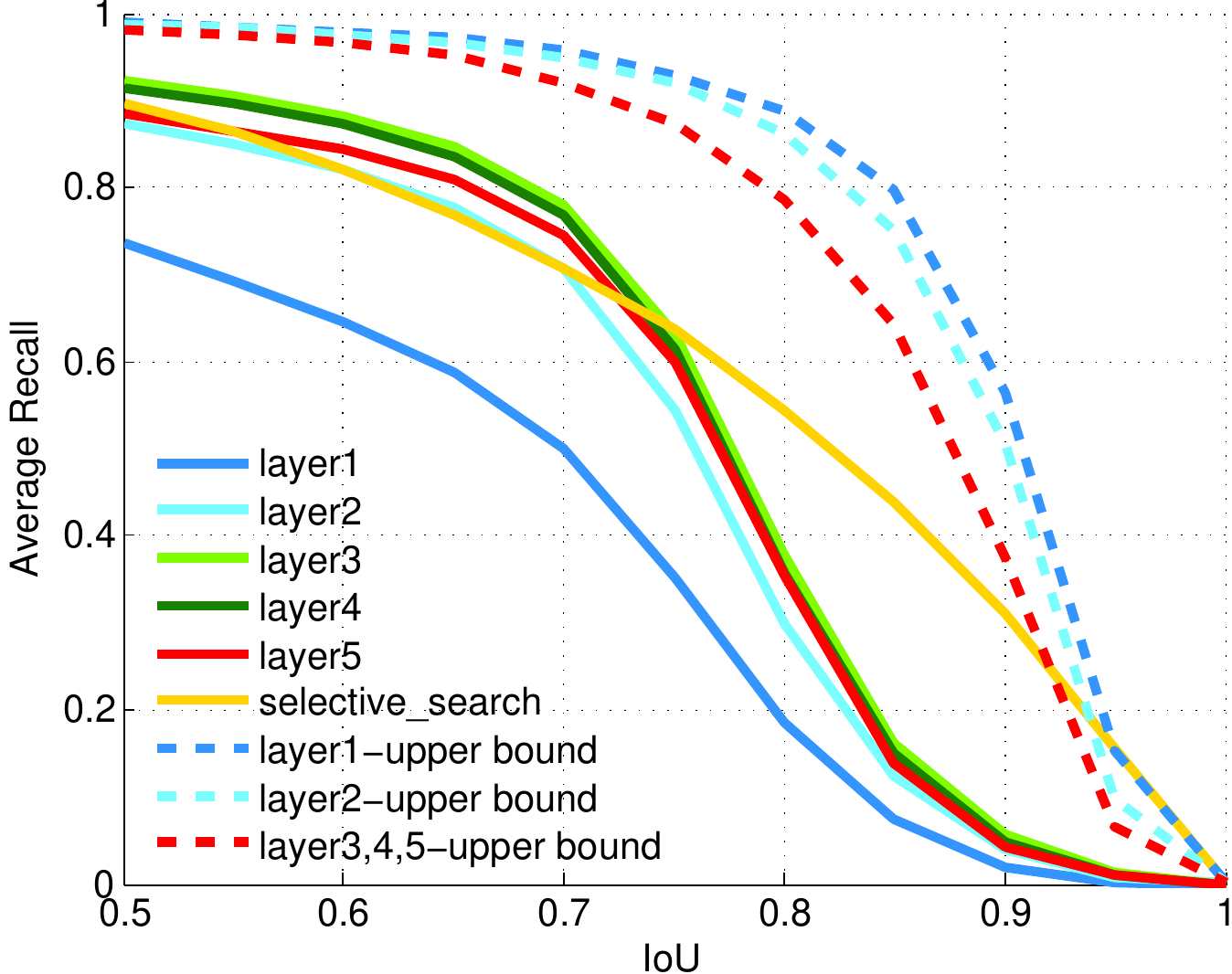}
&
\includegraphics[width=0.32\linewidth]{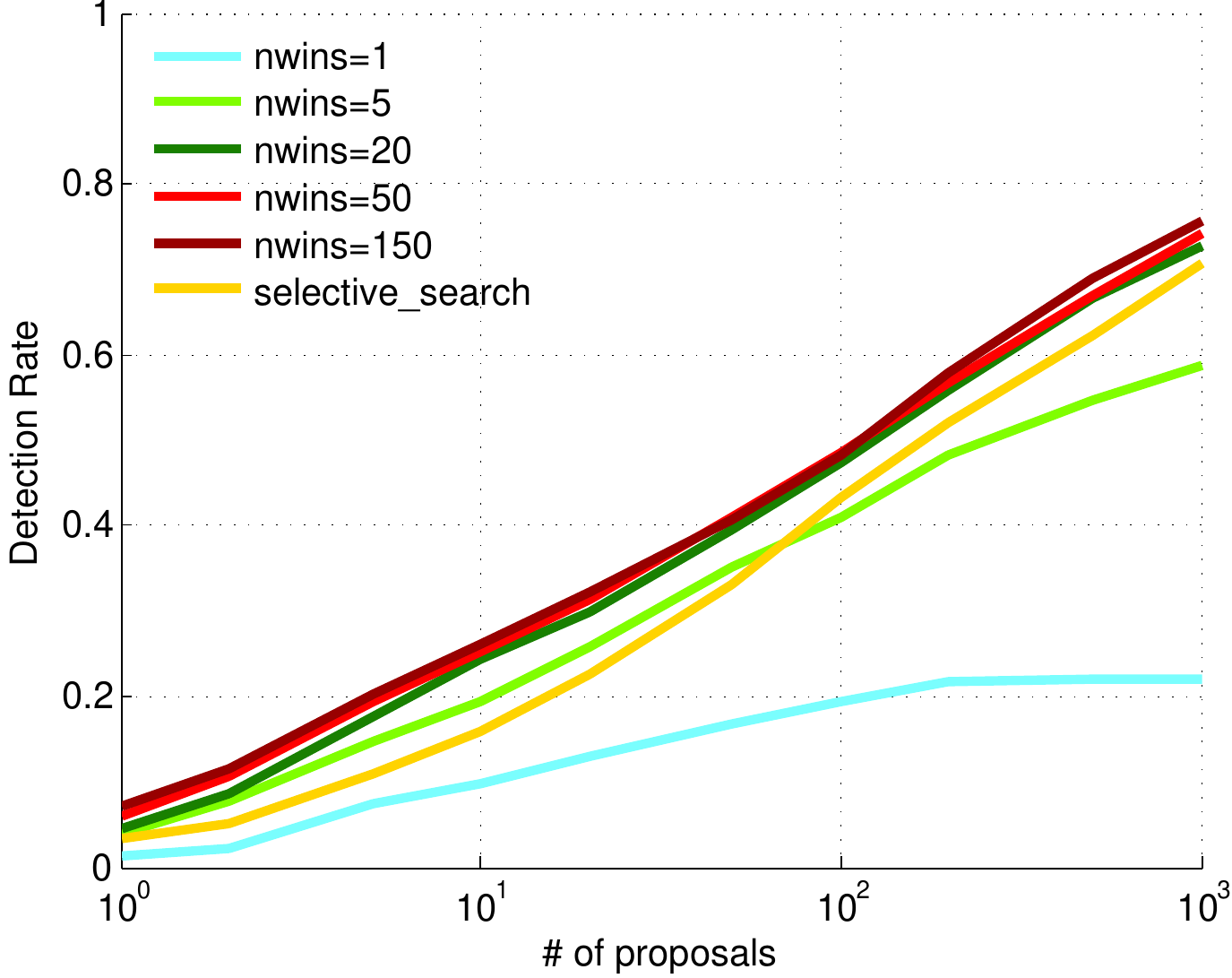}

%\end{center}
\end{tabular}
%}
\end{center}
\caption{({\bf{Left}}) Recall versus number of proposals for \texttt{IoU}=0.7. ({\bf{Middle}}) recall versus overlap for 1000 proposals for different layers. ({\bf{Right}}) Recall versus number of proposals at \texttt{IoU}=0.7 on layer 5 for different number of window sizes. All are reported on the PASCAL VOC 2007 test set.}
\label{fig:layer}
\end{figure*}

\begin{figure*}
\begin{center}
%\scalebox{0.8}
%{
\begin{tabular}{ccc}
%\begin{center}
\includegraphics[width=0.32\linewidth]{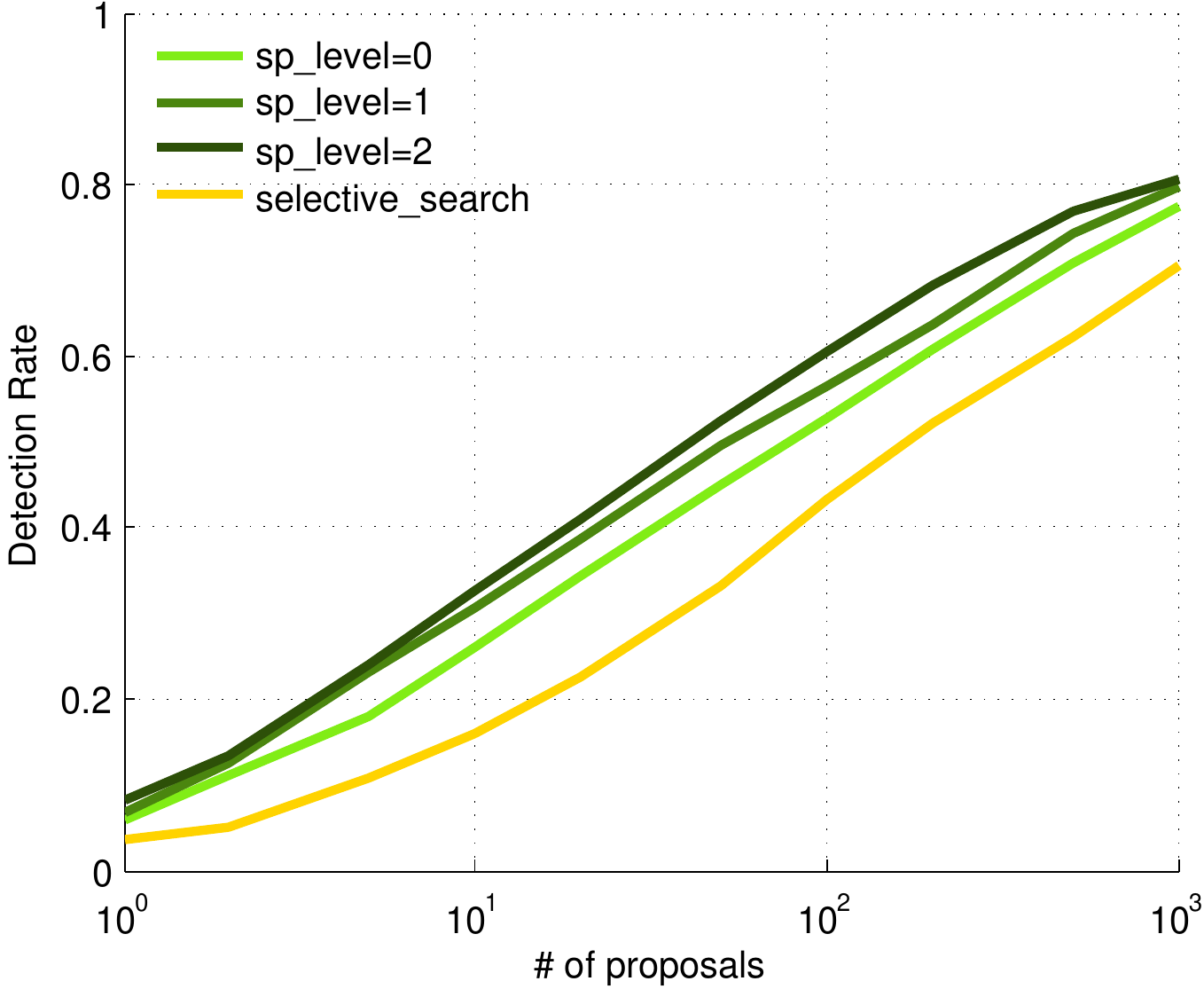}
&
\includegraphics[width=0.32\linewidth]{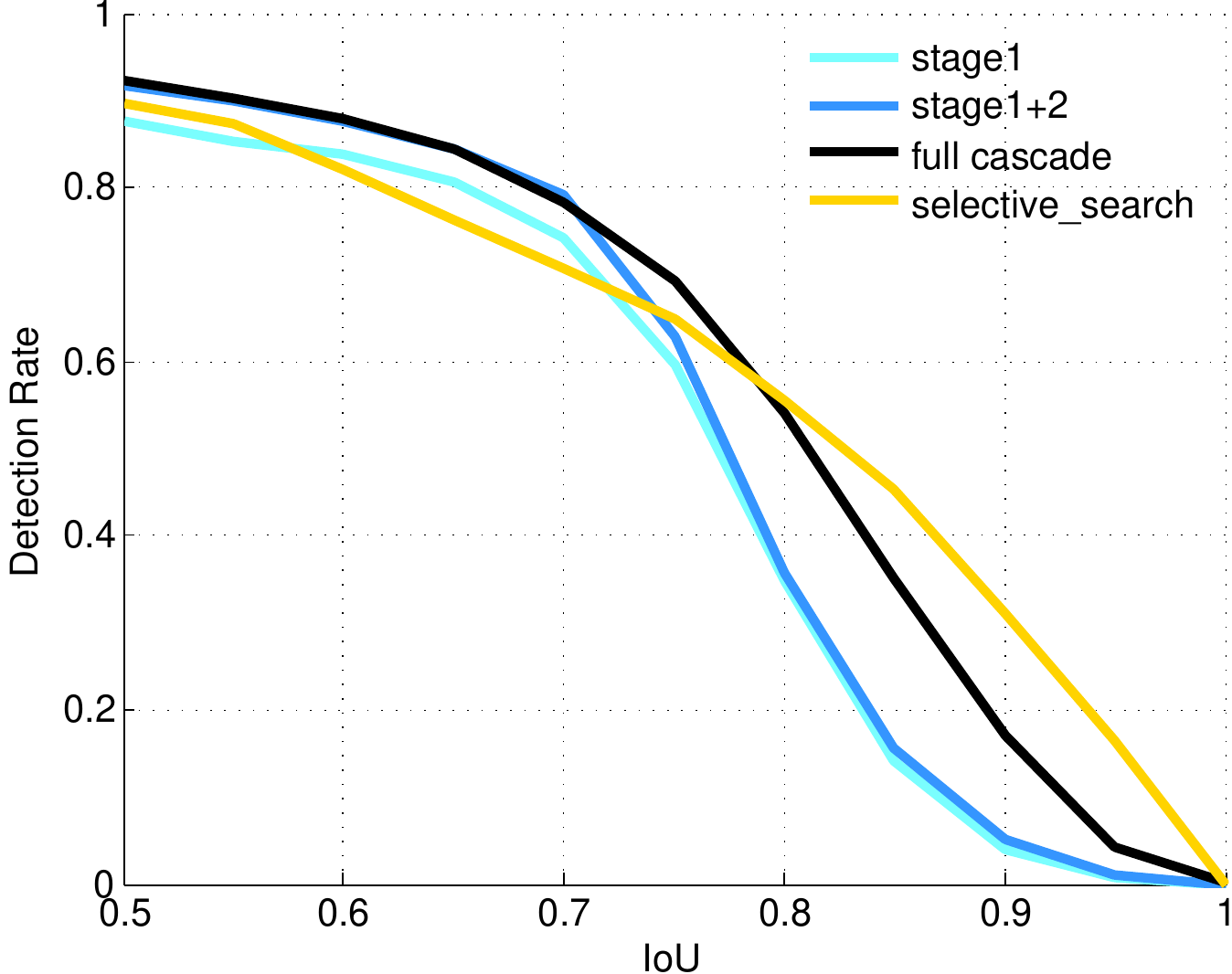}
&
\includegraphics[width=0.32\linewidth]{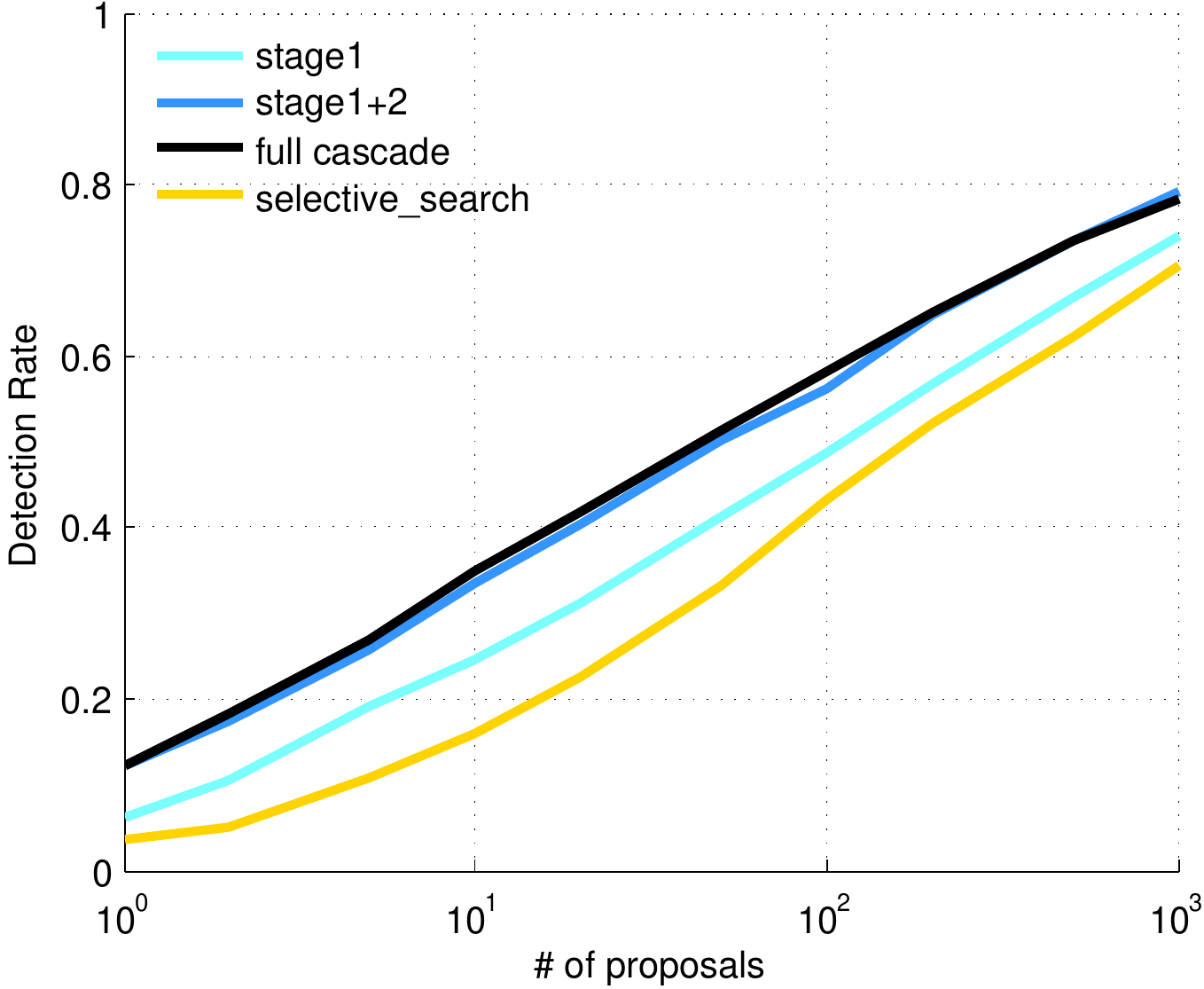}

%\end{center}
\end{tabular}
%}
\end{center}
\caption{({\bf{Left}}) Recall versus number of proposals in \texttt{IoU}=0.7 for different spatial pyramid levels ({\bf{Middle}}) Recall versus \texttt{IoU} for 1000 proposals for different stages of the cascade. ({\bf{Right}}) Recall versus number of proposals in \texttt{IoU}=0.7 for the different stages of the cascade. All are reported on the PASCAL VOC 2007 test set.}%MP:the caption should be updated}
\label{fig:var2}
\end{figure*}

%table of the cascade structure
% Stage Layer Pooling Pyramid-levels NMS Time
\begin{table*}[t]
\centering
%\scalebox{0.8}
%{
\begin{tabular}{cccccccc}%l|*{6}{c}}
	 Layer & Feature map size      & Recall(\#1000,0.5) & Max(0.5) & Recall(\#1000,0.8) & Max(0.8)\\
\hline
     5   &  $36\times52\times256$  & 88\% & 97\% & 36\%  & 70\%  \\
     4   &  $36\times52\times256$  & 91\% & 97\% & 36\%  & 79\%  \\
     3   &  $36\times52\times256$  & 92\% & 97\% & 38\%  & 79\%  \\
     2   &  $73\times105\times396$  & 87\% & 98\% & 29\%  & 86\%   \\
     1   &  $146\times210\times96$   & 73\% & 99\% & 18\%  & 89\%   \\
\end{tabular}
%}
\caption{Characteristics and performance of the CNN layers. Feature map size is reported for an image of size $600 \times 860$. Recall(\#1000,$\beta$) is the recall of 1000 proposals for the overlap threshold $\beta$. Max($\beta$) is the maximum recall for the overlap threshold $\beta$ using our selected window sizes set. }
\label{table:cnn_char}
\end{table*}

To evaluate the quality of our proposals we use the PASCAL VOC 2007 dataset~\citep{pascal}.
PASCAL VOC 2007 includes 9,963 images with 20 object categories. 4,952 images are used for testing, while the remaining ones are used for training. 
We use two different evaluation metrics; the first is Detection Rate (or Recall) vs. Number of proposals. This measure indicates how many objects can be recalled for a certain number of proposals.
We use Intersection over Union (\texttt{IoU}) as evaluation criterion for measuring the quality of an object proposal $\omega$. \texttt{IoU} is defined as $|\frac{\omega\cap b}{\omega\cup b}|$ where $b$ is the ground truth object bounding box. Initially, an object was considered correctly recalled if at least one generated window had an \texttt{IoU} of $0.5$ with it, the same overlap used for evaluating the detection performance of a method. Unfortunately this measure is too lose because a detector, for working properly, needs also good alignment with the object \citep{Rodrigo14}. Thus we evaluate our method for an overlap of $0.7$ as well. 
We also evaluate recall vs. overlap for a fixed number of proposals. As shown in \citet{Rodrigo14}, the average recall obtained from this curve seems highly correlated with the performance of an object detector built on top of these proposals. % using the corresponding generating method.

In this section, we investigate the effect of different parameters of our method, namely the different convolutional layers, the number of used window sizes and different levels of spatial pyramid pooling. %AG:pyramids,pooling,...

\paragraph{\textbf{Layers}}
We evaluate each convolutional layer (from 1 to 5) of Alexnet \citep{alexnet} using the sliding window settings explained above. %We use Alexnet which is trained by Caffe toolbox \citet{caffe}. 
For sake of simplicity, we do not add spatial pyramids on top of pooled features in this set of experiments. As shown in Fig.~\ref{fig:layer}~({\bf left}) the top convolutional layers of the CNN perform better than the bottom ones. Also their computational cost is lower as their representation is coarser. Note this simple approach already performs on par or even better than the best proposal generator approaches from the literature. For instance, our approach at layer 3 for 100 proposals achieves a recall of $52\%$, whereas
selective search \citep{selectivesearch} obtains only $40\%$. This makes sense because the CNN features are specific for object classification and therefore can easily localize the object of interest.

However, this is only one side of the coin. If we compare the performance of the CNN layers for high overlap (see Fig.~\ref{fig:layer}~({\bf middle})), we see that segmentation-based methods are much better \citep{selectivesearch, MCG}. For instance the recall of selective search for 1000 proposals at $0.8$ overlap is around $55\%$ whereas ours at layer 3 is only $38\%$. This is due to the coarseness of the CNN feature maps that do not allow a precise bounding box alignment to the object. In contrast, lower levels of the net have a much finer resolution that can help to align better, but their encoding is not powerful enough to properly localize objects. In Fig.~\ref{fig:layer}~({\bf middle}) we also show the maximum recall for different overlap that a certain layer can attain with our selected sliding windows. In this case, the first layers of the net can recall many more objects with high overlap. This shows  that a problem of the higher layers of the CNN is the lack of a good spatial resolution. 

In this sense we could try to change the structure of the net in a way that the top layers still have high spatial resolution. However, this would be computationally expensive and, more importantly, it would not allow to reuse the same features used for detection. Instead, in the next section we propose an efficient way to leverage the expressiveness of the top layers of the net together with the better spatial resolution of the bottom layers.

\paragraph{\textbf{Number of Window Sizes}}
In Fig.~\ref{fig:layer}~({\bf right}) we present the effect of a varying number of window sizes in the sliding window procedure for proposal generation. The windows are selected based on the greedy algorithm explained in Sec~\ref{subsec:basic}. 
As the number of used window sizes increases, we obtain a better recall at a price of a higher cost.
In the following experiments we will fix the number of windows to $50$ because that is a good trade-off between speed and top performance.
The values in the figure refer to layer 5, however, similar behavior has been observed for the other layers as well.

\paragraph{\textbf{Spatial Pyramid}}
We evaluate the effect of using a spatial pyramid pooling in Fig.~\ref{fig:var2}~({\bf left}). As expected, adding geometry improves the quality of the proposals. Moving from a pure average pooling representation (sp\_level=0) to a $2\times2$ pyramid (sp\_level=1) gives a gain that varies between $2$ and $4$ percent in terms of recall, depending on the number of proposals. Moving from the $2\times2$ pyramid to the $4\times4$ (sp\_level=2) gives a slightly lower gain. At $4\times4$ the gain does not saturate yet. However, as we aim at a fast approach, we also need to consider the computational cost, which is linear in the number of spatial bins used. Thus, the representation of a window with a $2\times2$ spatial pyramid is 5 times slower than a flat representation and the $4\times4$ pyramid is $21$ times slower. For this reason, in our final representation we limit the use of the spatial pyramid to $2\times2$ spatial pyramid. 

%\paragraph{\textbf{Scales}}
%Another set of experiments of variation is about the number of image scales. Based on experiments that their results are in Figure ?, with more scales of image, \methodname~can use more models of objectness which are trained separately on each scale, So it can use more benefits of all scales. [224, 300, 400, 600] are scales which used to extract CNN features. 

%------------------------------------------------------------------------
% Coarse-to-fine Inverse Cascade
%------------------------------------------------------------------------
\section{Coarse-to-fine Inverse Cascade}
\label{sec:hierarchy}

Even if the features used for our window proposals come without any additional computational cost (because they are needed for the detector), still a dense evaluation in a sliding window fashion over the different layers would be too expensive.
Instead here we leverage the structure of the CNN layers to obtain a method that combines in an efficient way the high recall of the top convolutional layers of a CNN, with the fine localization provided at the bottom layers of the net. In Table~\ref{table:cnn_char} we summarize the characteristics of each CNN layer.

We start the search with the top convolutional layers of the net, that have features well adapted to recognize high-level concepts like objects and actions, but are coarse, and then move to the bottom layers, that use simpler features but have a much finer spatial representation of the image (see Fig.~\ref{fig:overview}). As we go from a coarse to a fine representation of the image and we follow a flow that is exactly the opposite of how those features are computed we call this approach \textit{coarse-to-fine inverse cascade}.
We found that a cascade with 3 layers is an optimal trade-off between complexity of the method and gain obtained from the cascading strategy.

\paragraph{\textbf{Stage 1: Dense Sliding Window on Layer 5}}
The first stage of the cascade uses layer 5. As the feature representation is coarse, we can afford a dense sliding window approach with $50$ different window sizes collected as explained in Sec.~\ref{subsec:basic}. Even though a pyramid representation could further boost the performance, we do not use spatial binning at this stage to not increase the computational cost. We linearly re-scale the window scores to $[0,1]$ such that the lowest and highest scores are mapped to $0$ and $1$ respectively. Afterwards we select the best $N_1=4000$ windows obtained from a non-maximum suppression algorithm with threshold $\beta+0.05$ in order to propagate them to the next stage. %MP:to select a fixed number of proposals it's not enough the NMS, it's a bit misleading... how do you map from 0 to 1?

\paragraph{\textbf{Stage 2: Re-scoring Selected Windows on Layer 3}}
In this stage, as we use a reduced set of windows, we can afford to spend more computation time per window. Therefore we add more geometry in the representation by encoding each window with a pyramid representation composed of two levels: $1\times1$ and $2\times2$. The proposal scores from this layer are again mapped to $[0,1]$. The final score for each proposal is obtained  by multiplying the scores of both stages. Afterwards we apply a non-maximum suppression with overlap threshold $\beta+0.05$ and select the $3000$ best candidates. At the end of this stage, we aggregate the boxes from different scales using non-maximum suppression with threshold $\beta$ and select the $N_{desired}=1000$ best for refinement.

\paragraph{\textbf{Stage 3: Local Refinement on Layer 2}}
The main objective of this final stage is to refine the localization obtained from the previous stage of the cascade. For this stage the best candidate is layer 2 because it has a higher resolution than upper layers and contains low-level information which is suitable for the refinement task. Specifically, we refine the $N_{desired}$ windows received from the previous stage using the procedure explained in~\citet{edgebox}. To this end, we train a structured random forest~\citep{structurededge} on the second layer of the convolutional features to estimate contours similarly to DeepContour~\citep{deepcontour}. After computing the edgemap, a greedy iterative search tries to maximize the score of a proposal over different locations and aspect ratios using the scoring function used in~\cite{edgebox}. %MP: we should explain it instead of saying that is like in edgebox... 
It is worth mentioning that since our contour detector is based on the CNN-features, we again do not need to extract any extra features for this step.

\subsection{Evaluation for object proposals}
\label{subsec:eval_hierarchy}
%table of the cascade structure
% Stage Layer Pooling Pyramid-levels NMS Time
\begin{table*}[t]
\centering
\begin{tabular}{cccccccc}%l|*{6}{c}}
	Stage & Layer & input candidates & Method       & Pyramid & NMS    & Total time per image\\
\hline
     1    &   5   &  $\sim$80.000  	& Slid. Window 	& $1$       & Yes &  0.30s\\
     2    &   3   &  4.000  		& Re-scoring    & $1+2\times2$       & Yes &  0.25s\\
     3    &   2   &  1.000  		& Refinement    & -       & No  &  0.20s \\
\end{tabular}
\caption{Characteristics of the stages of our inverse cascade (NMS: non maximum suppression).}
\label{table:stages}
\end{table*}
We discuss the performance of the inverse cascade stage by stage in terms of both computational cost and performance. A summary of the computational cost of each stage is given in Table~\ref{table:stages}.
The entire cascade has a computational cost of $0.75$ 
on a 8-core CPU of 3.50GHz, which is the composition of $0.3$ , $0.25$ and $0.2$ for the first, second and third stage respectively. Note the first stage is very fast because even if we use a dense sliding window approach, with the integral image and without any pyramid level the cost of evaluating each window is very low. 

As shown in Fig.~\ref{fig:var2}~({\bf middle} and {\bf right}), 
%the first two stages contribute to improve different parts of the curve. 
the second stage is complementary to the first and employed with a $2\times2$ pyramid improves the recall of the cascade by $5\%$. However, this boost is valid only up to an overlap of $0.75$. After this point the contribution of the second stage is negligible. This is due to the coarse resolution of layer $5$ and $3$ that do not allow a precise overlap of the candidate windows with the ground truth object bounding boxes. We found that, for our task, layer 3 and 4 have a very similar performance (Recall@1000 is $79\%$ in both cases) and adding the latter in the pipeline did not help in improving performance (Recall@1000 is still $79\%$).

As shown in \citet{Rodrigo14}, for a good detection performance, not only the recall is important, but also a good alignment of the candidates is needed. At stage $3$ we improve the alignment without performing any further selection of windows; instead we refine the proposals generated by the previous stages by aligning them to the edges of the object. In our experiments for contour detection we observed that the first layer of the CNN did not provide as good performance as layer 2 (0.61 vs. 0.72 AP on BSDS dataset~\citep{arbelaez2011contour}), so we choose the second layer of the network for this task. Fig.~\ref{fig:var2}~({\bf middle}) shows that this indeed improves the recall for high  \texttt{IoU} values (above 0.7).

%------------------------------------------------------------------------
% Proposals in Videos
%------------------------------------------------------------------------
\section{Proposals in Videos}
\label{sec:time_cont}
Given a video sequence of length $T$, the goal is to generate a set of action proposals (tubes). Each proposal $P=\left \{R_1,...,R_t,...,R_{T}\right \}$ corresponds to a path from the box $R_1$ in the first frame to the box $R_{T}$ in the last frame, and it spatially localizes the action. 

In case of object proposals, applying the inverse cascade to an image already gives the desired output. When the goal is to find proposals in videos, we need a) to capture the motion information that a video naturally provides us with, and b) to satisfy time continuity constraints over time. 

One advantage of \methodname~is that it can be setup on top of any fine-to-coarse convolutional network regardless of its input/output (and possibly its architecture). 
To benefit from both appearance and motion cues in a given video, we use two networks for the task of action proposal generation. The first network takes as input the RGB frames of a video, %operates on appearance cues and capture the appearance of the actors, objects and scene in each frame of the video. This network has 
and is based on an Alexnet-like architecture, fine-tuned on VOC2007 for the task of object detection. The second network takes as input the optical flow of each frame extracted from the video. %and captures motion cues of actors and objects in each frame. 
We use the motion-CNN network of~\citet{gkioxari2015finding} trained on UCF101 (split1)~\citep{soomro2012ucf101}. The architecture of this network is identical to the first network.  

To generate a set of proposals in each frame, in the first and second stage of \methodname, we use an early fusion strategy, concatenating %MP: is it correct? 
the feature maps generated by both networks and treating them as a single set of feature maps. For the last stage, since it is an alignment process, we only use the feature map of the appearance network. 

So far the output is a set of proposals in each frame. In order to make  the proposals temporally coherent, we follow the procedure of~\citet{gkioxari2015finding} and link the proposals of each single frame over time into tubes. We define the linking scoring function between every two consecutive boxes $R_t$ and $R_{t+1}$ as follows:
$$S(R_t,R_{t+1}) = C(R_t)+C(R_{t+1})+O(R_t,R{t+1})$$
where $C(.)$ is the confidence score for a box and $O(.)$ is the intersection over union value if the overlap of the two boxes is more than $0.5$, otherwise it is $-Inf$. Intuitively, the scoring function  gives a high score if the two boxes $R_t$ and $R_{t+1}$ overlap significantly and if each of them most likely contains an object of an action. 

Finally, we are interested in finding the optimal path over all frames. To this end, we first compute the overall score for each path $P$ by $\sum_{t=1}^{T-1} \sum_{i,j \in P} S(R_t^i,R_{t+1}^j)$. Computing score for all the possible paths can be done efficiently using the Viterbi algorithm. The optimal path $P^{opt}$ is then the one with highest score. After finding the best path, all boxes in $P^{opt}$ are removed and we solve the optimization again in order to find second best path. This procedure is repeated until finding the last feasible path (those paths whose scores are higher than $-Inf$). We consider each of these paths as an action proposal.

\subsection{Evaluation for action proposals}
\label{subsec:eval_hierarchy_a}
We evaluate the performance of the inverse cascade for action proposals on the UCF-Sports~\citep{rodriguez2008action} dataset. We train our models on a training set that contains a total of $6,604$ frames. Additionally, we have $2,976$ frames in the test set, spread over $47$ videos. 

Like~\citet{van2015apt}, we measure the overlap between an action proposal and the ground-truth video tubes using the average intersection-over-union score of 2D boxes for all frames where there is either a ground-truth box or proposal box. Formally:
$$Ovr(P,G)=\frac{1}{\left | F \right |}\sum_{t\in F}\frac{P_t\cap G_t}{P_t\cup G_t},$$
where $P$ and $G$ are action proposal and ground-truth action tube respectively. $F$ is the set of frames where either P or G is not empty. $G_t$ is empty if there is no action in the frame $t$ of the ground-truth tube. In this case, $P_t\cap G_t$ is set to $0$ for that frame.

Considering each frame as an image and applying \methodname~on each frame individually, the frame-based recall of objects/actors for an \texttt{IoU} of $0.7$ is $78\%$ for 10 windows and $96\%$ for $100$ windows. One possible explanation for such a promising frame-based recall is that an action mainly contains an actor performing it and hunting that actor in each frame is relatively easier than hunting general objects in the task of object proposal generation. However, this does not take into account any temporal continuity. Constructing tubes from these static windows, which results in our action proposals, is our final goal. 

\begin{figure*}
\begin{center}
\scalebox{0.93}
{
\begin{tabular}{ccc}
%\begin{center}
\includegraphics[width=0.32\linewidth]{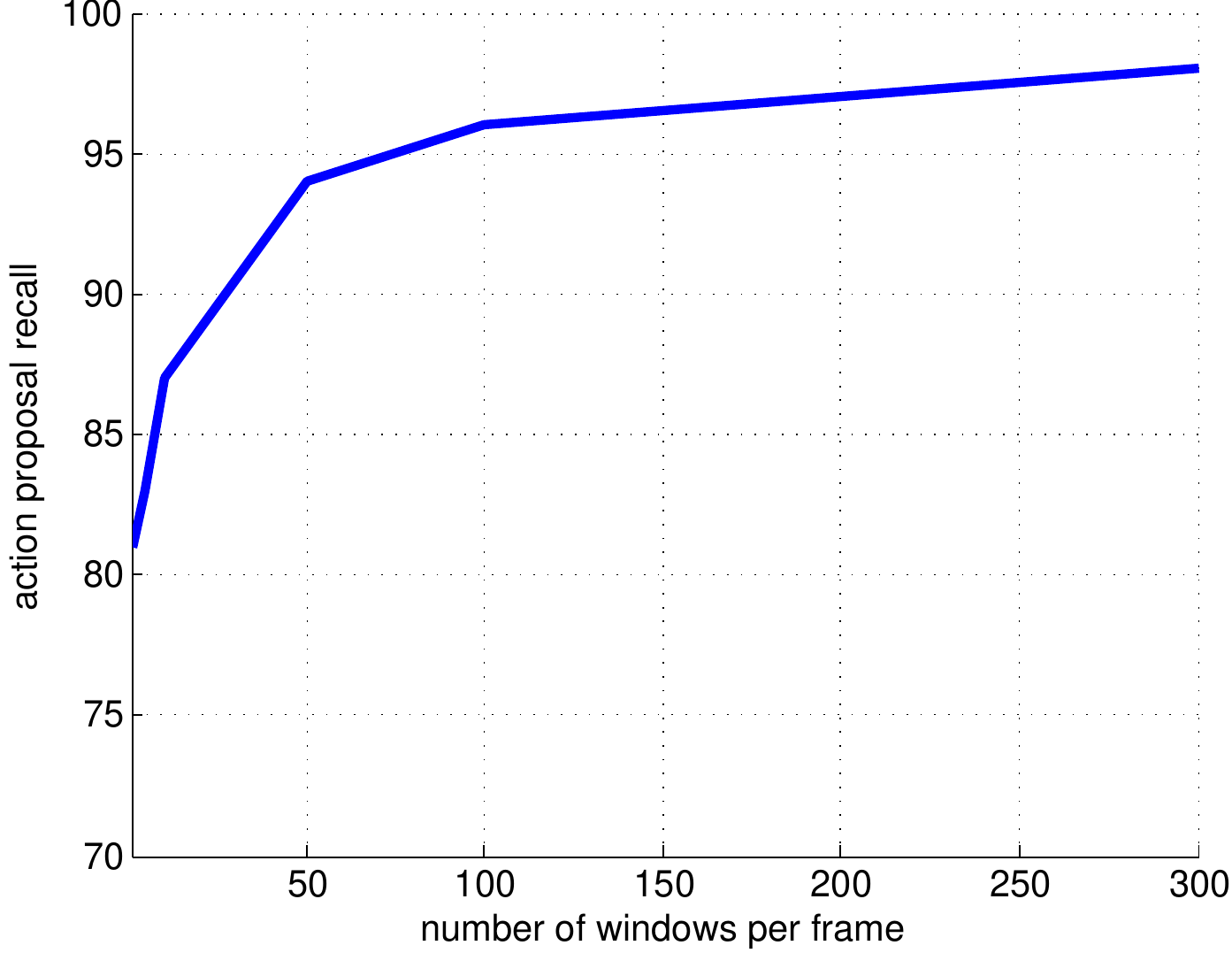}
&
\includegraphics[width=0.32\linewidth]{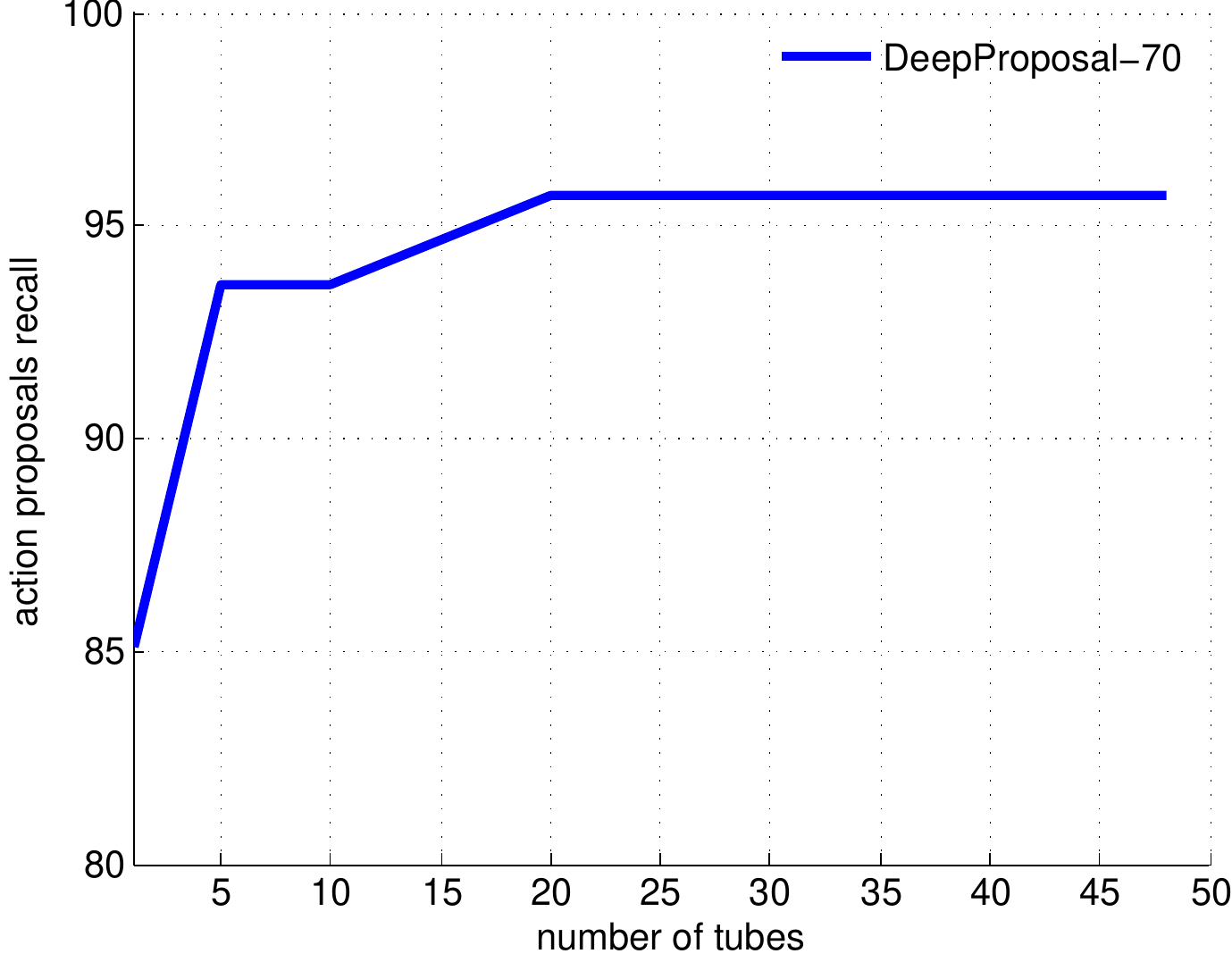}
&
\includegraphics[width=0.32\linewidth]{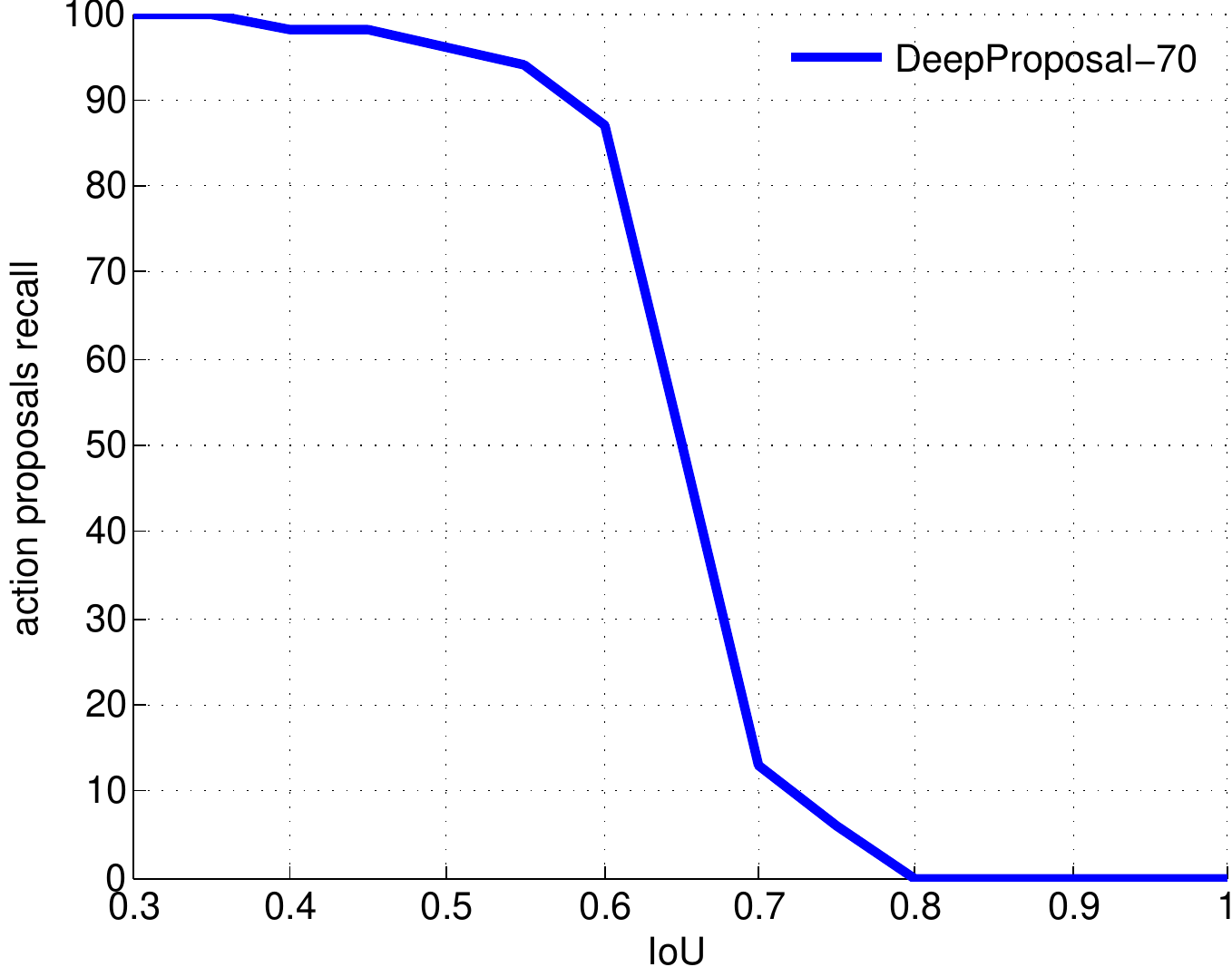}
%\end{center}
\end{tabular}
}
\end{center}
\caption{({\bf{left}}) Action proposals recall at IoU of $0.5$ varying the number of windows per frame. ({\bf{middle}}) Recall versus number of proposals in \texttt{IoU}=0.5 for the different number of action proposals. ({\bf{right}}) Action proposal recall at different IoU values for 20 action proposals. All results are reported on the UCF-sports test set.}%MP:the caption should be updated}
\label{fig:action_exp}
\end{figure*}

The extension of the inverse cascade for actions introduces an additional parameter which is the number of windows that we select in each frame. In figure~\ref{fig:action_exp} ({\bf{left}}) we show the recall of the action proposals while varying the number of windows we select per frame. 
As expected, selecting more windows in each frame leads to a higher recall of the action proposals. However, it also leads to an increasing computational cost, since the computational complexity of the Viterbi algorithm is proportional to the square of the number of windows per frame. For example, the Viterbi algorithm for a video of length $240$ frames takes $1.3$ and $12.1$ seconds for $N=100$ and $N=300$ respectively.
From now on, during all the following experiments we select $N=100$ windows per frame to have a good balance between performance and time complexity.

Figure~\ref{fig:action_exp} ({\bf{middle}}) shows the action proposals recall for different number of proposals. As it is shown even for a very small number of proposals, \methodname~obtains very good performance as already observed also for object proposals.
Figure~\ref{fig:action_exp} ({\bf{right}}) shows the recall of our method for 20 action proposals (tubes) per video over different IoU values. Our method works very well in the regime of $[0.3..0.6]$. Notice that the definition of action proposals recall is different than object proposals recall and the performance in \texttt{IoU}=$0.5$ is already quite promising.

%------------------------------------------------------------------------
% Experiments
%------------------------------------------------------------------------
\section{Experiments}
\label{sec:comparison}
In this section we compare the quality of the proposed \methodname~with state-of-the-art object (section~\ref{subsec:objectp}) and action (section~\ref{subsec:actionp}) proposals. 

\subsection{Object Proposals}
\label{subsec:objectp}
To evaluate our proposals, like previous works on object proposal generation, we focus on the well-known PASCAL VOC 2007 dataset. PASCAL VOC 2007 \citep{pascal} includes $9,963$ images divided in 20 object categories. $4,952$ images are used for testing, while the remaining ones are used for training. 
We compare the quality of our \methodname~in terms of recall and localization accuracy with multiple state-of-the-art methods. % in section~\ref{subsubsec:results}. Then, in section~\ref{subsubsec:detectionRes} 
Detection results and run-time are reported for PASCAL VOC 2007 \citep{pascal}, integrating \methodname~in the Fast-RCNN framework~\citep{girshick15fastrcnn}. %In section~\ref{subsubsec:generalization}, 
Then, we evaluate the generalization performance of \methodname~on unseen categories and finally some qualitative comparisons are presented. % in section~\ref{subsubsec:qualitative}.

\begin{figure*}[t]
\begin{center}
%\scalebox{0.9}
%{
\begin{tabular}{c@{}c@{}c@{}c}
\includegraphics[width=0.32\linewidth]{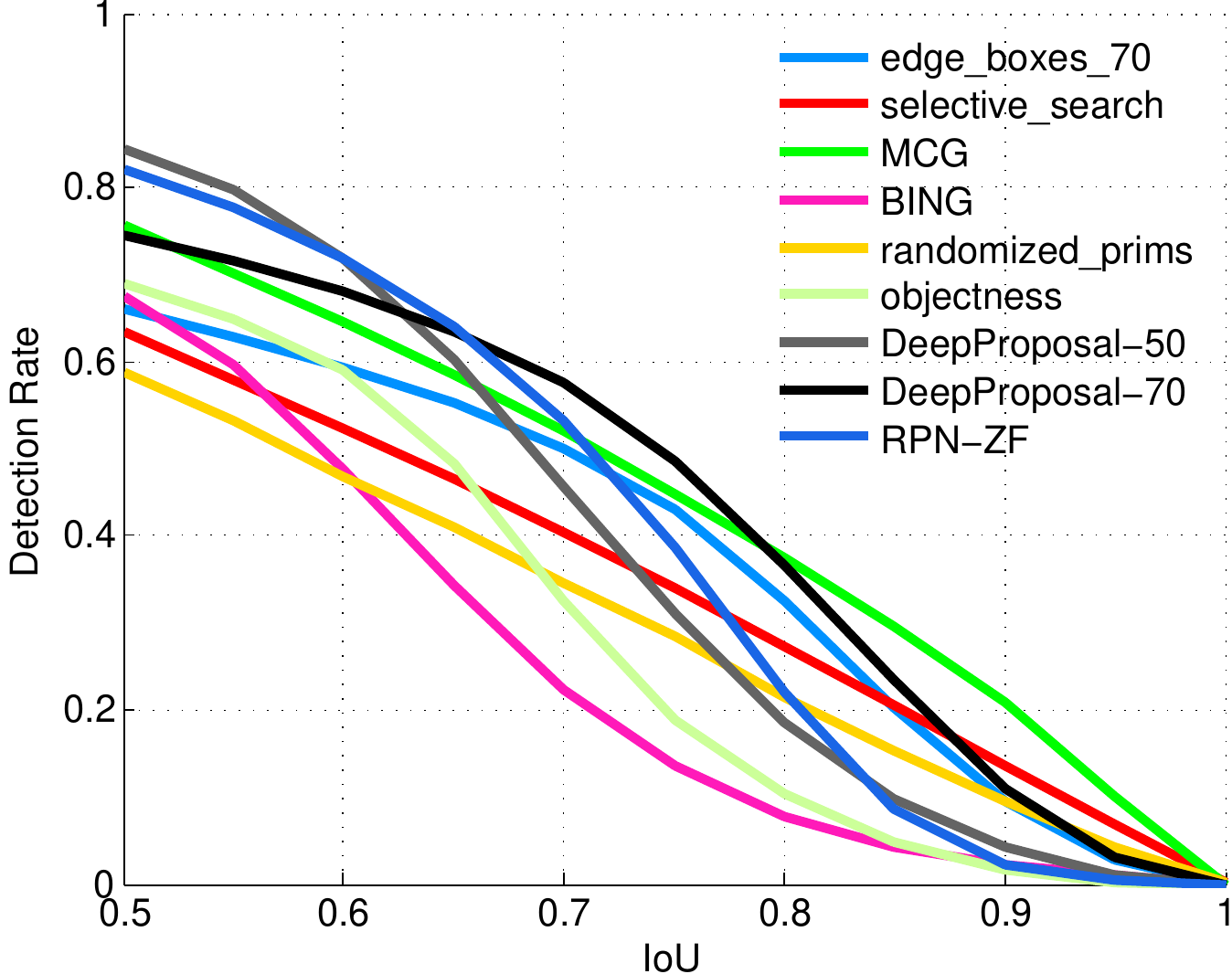}
&
\includegraphics[width=0.32\linewidth]{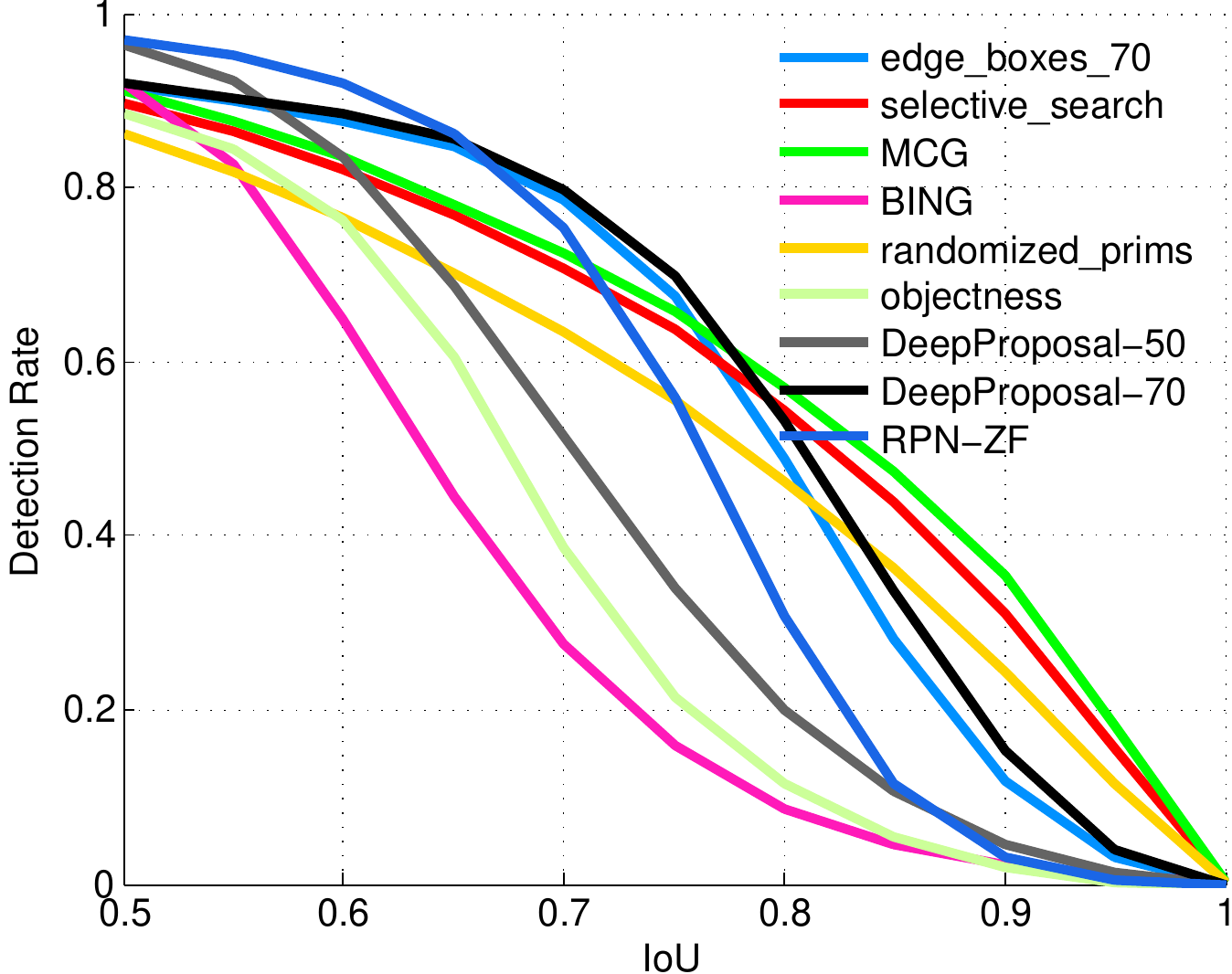}
&
\includegraphics[width=0.32\linewidth]{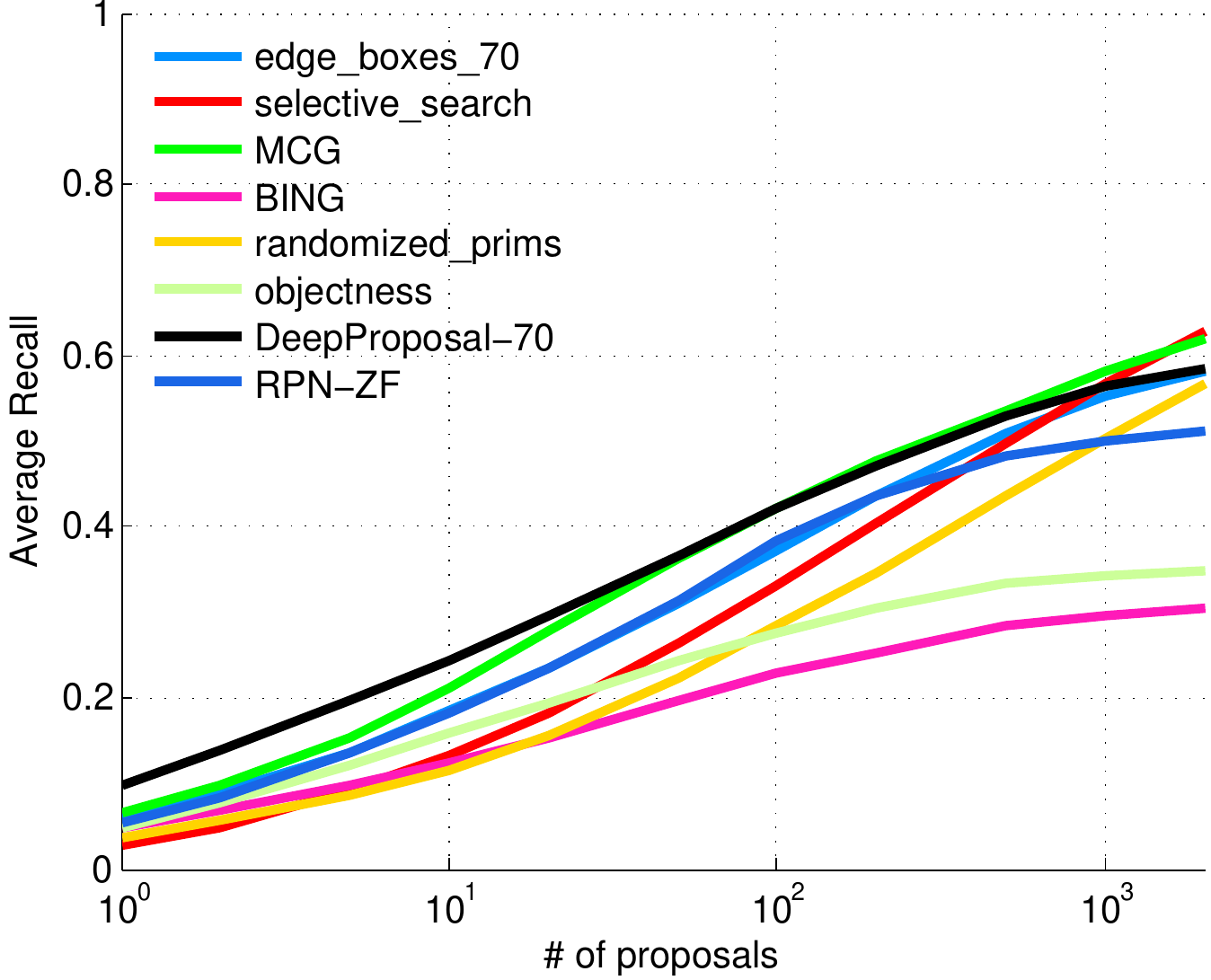}
\end{tabular}
%}
%\vspace{-0.5cm}
\caption{Recall versus IoU threshold on the PASCAL VOC 2007 test set for ({\bf{left}}) 100 proposal windows and ({\bf{middle}})1000 proposal windows. ({\bf{right}}) Average Recall between [0.5,1] \texttt{IoU} on the PASCAL VOC 2007 test set.}
\label{fig:iou}
\end{center}
\end{figure*}

%------------------------------------------------------------------------
\paragraph{\textbf{Comparison with state-of-the-art}}
%\subsubsection{Comparison with state-of-the-art}
%\label{subsubsec:results}
%------------------------------------------------------------------------
We compare our \methodname~against well-known, state-of-the-art object proposal methods. Fig.~\ref{fig:nprop} and Fig.~\ref{fig:iou} show the recall with a varying number of object proposals or \texttt{IoU} threshold respectively. %These curves reveal how \methodname~performs on varying \texttt{IoU}. 
From Fig.~\ref{fig:nprop}, we can see that, even with a small number of windows, \methodname~can achieve higher recall for any \texttt{IoU} threshold.
%EdgeBoxes \citet{edgebox} a recently published method, is the state-of-the-art in generating object proposal and has competitive accuracy in comparison with other methods like selective search \citet{selectivesearch} and MCG\citet{MCG} cross all \texttt{IoU} specially with lower number of windows. 
%We can compare methods based on recall precision and limitation of boxes that they're generating. 
%For example CPMC \citet{CPMC} has high quality object proposals but can not generate
%large number of them, so recall will decrease under this circumstance.sub
Methods like BING \citep{BING} and objectness \cite{objectness} are  providing high recall only at \texttt{IoU} = 0.5 because they are tuned for \texttt{IoU} of 0.5. % based on similar threshold for detection true positive. 

\begin{figure*}[t]
\begin{minipage}{.67\textwidth}
\begin{center}
\begin{tabular}{c@{}c@{}c@{}c}
\includegraphics[width=0.46\linewidth]{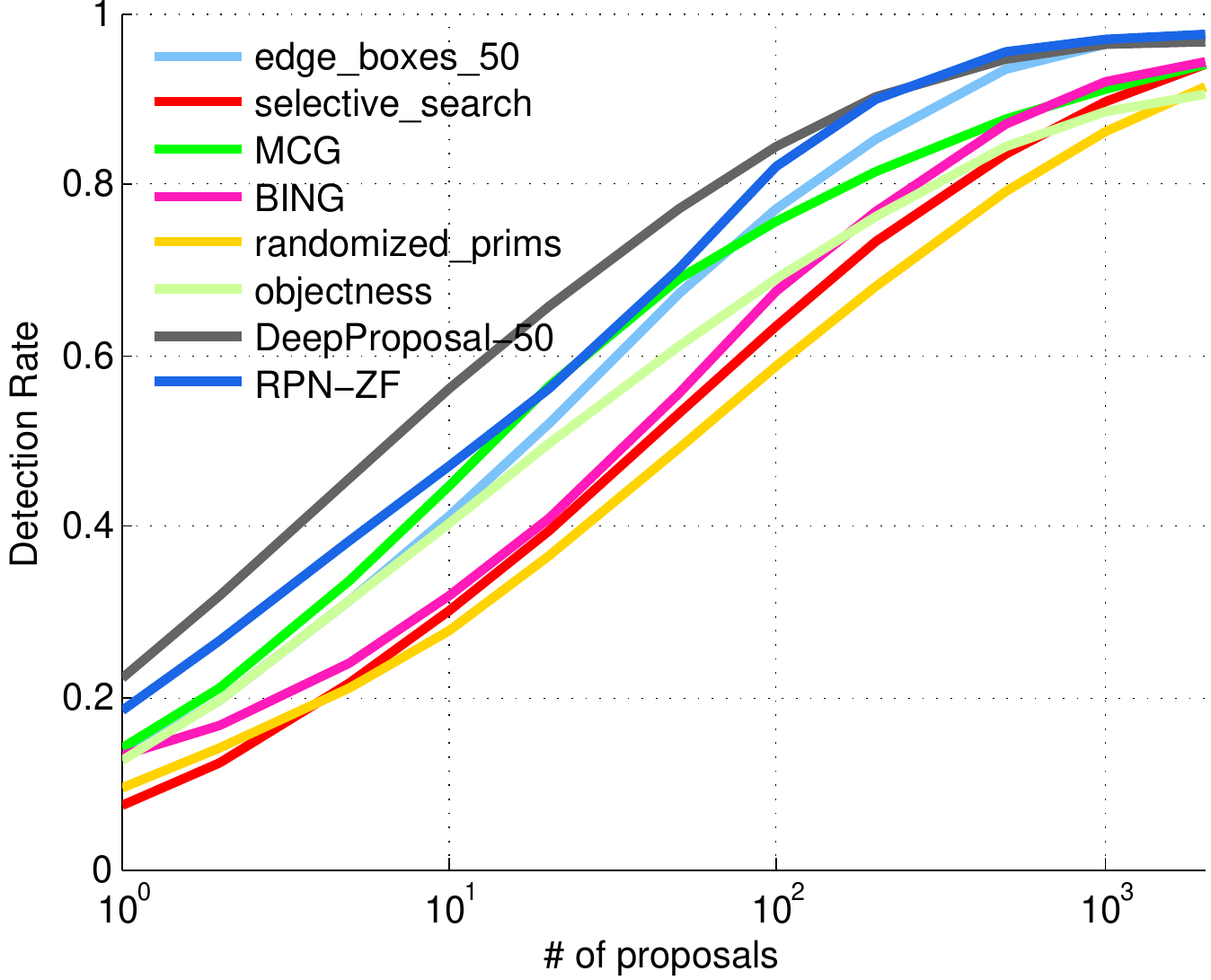}
&
\includegraphics[width=0.46\linewidth]{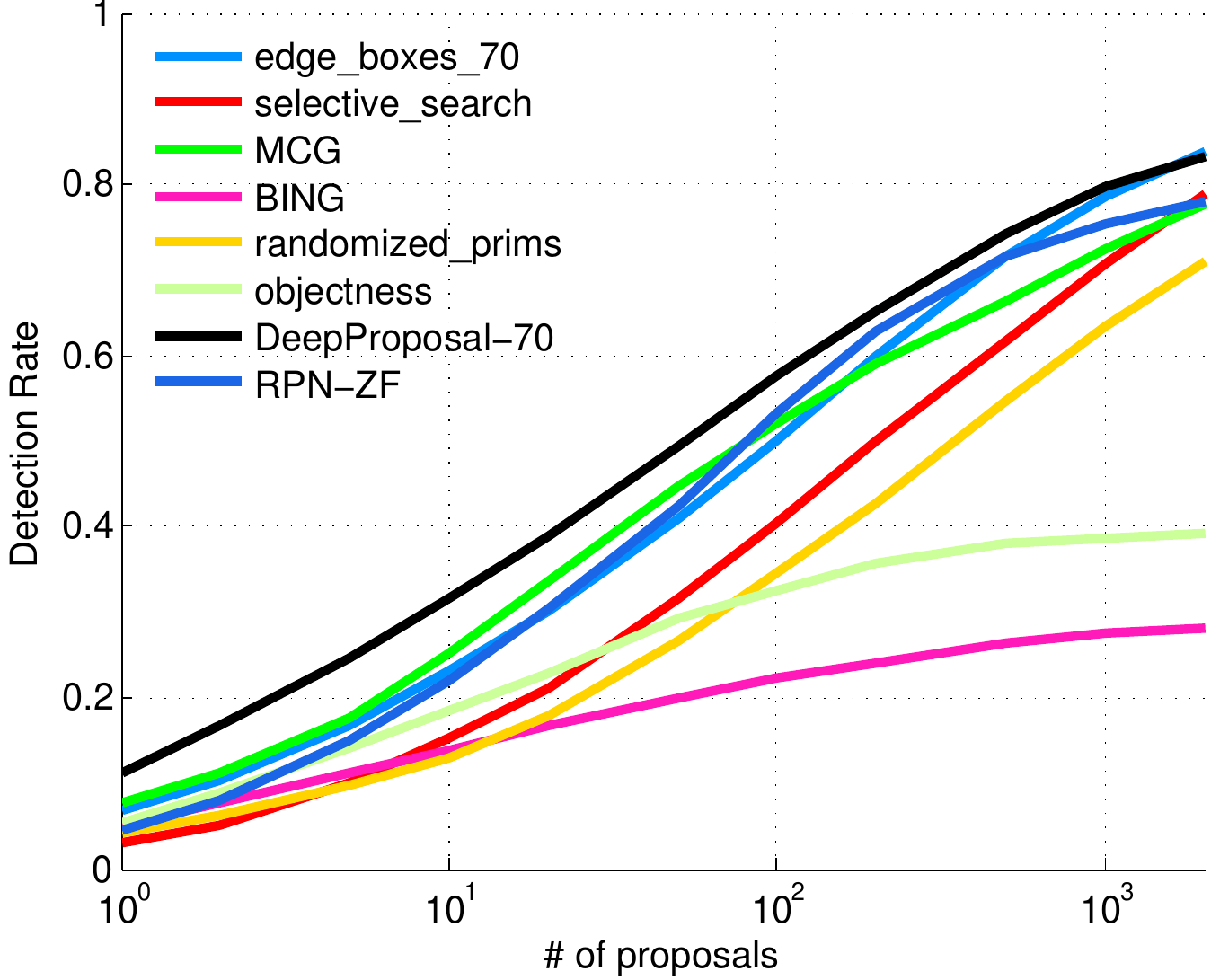}
\end{tabular}
%\vspace{-0.5cm}
\captionof{figure}{Recall versus number of proposals on the PASCAL VOC 2007 test set for ({\bf{left}}) IoU threshold 0.5 and ({\bf{right}})IoU threshold 0.7.}
\label{fig:nprop}
\end{center}
\end{minipage}%
\begin{minipage}{.31\textwidth}
\begin{center}
%\scalebox{0.6}
{
{\includegraphics[width=0.95\linewidth]{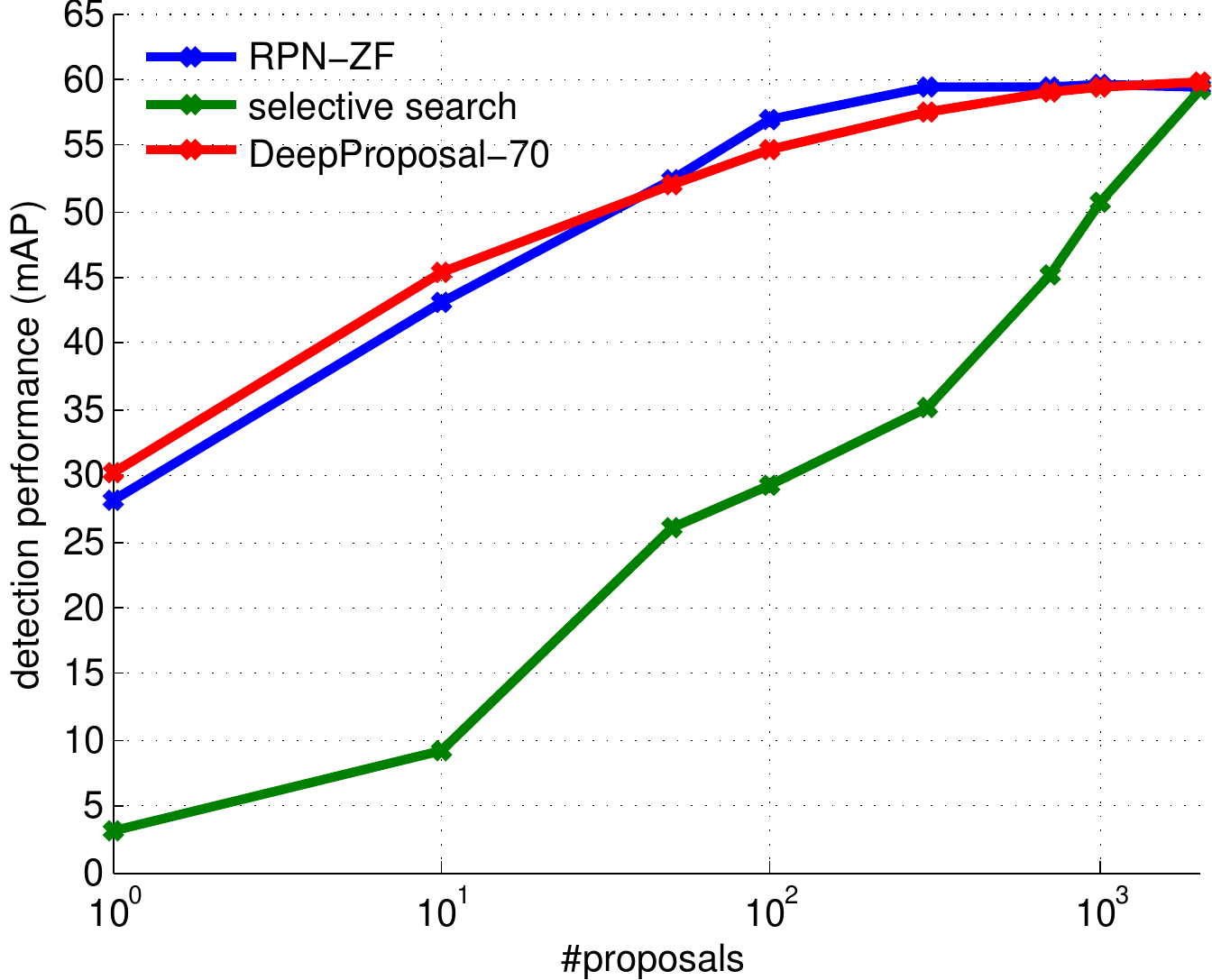}} %\cfbox{red}
}
\end{center}
\vspace{-0.2cm}
\captionof{figure}{Detection results on PASCAL VOC 2007.}
\label{fig:det_res}
\end{minipage}%
\end{figure*}

%When comparing the recall of several methods while varying the \texttt{IoU} thresholds (Fig.~\ref{fig:iou} {\bf{left}} and {\bf{middle}}), we observe that \methodname~achieve competitive or higher recall. 
In Table~\ref{table:cmpr} we summarize the quality of the proposals generated by the most promising methods. 
Achieving 75\% recall with \texttt{IoU} of $0.7$ would be possible with $540$ proposals of \methodname, $800$ of Edge boxes, $922$ of RPN-ZF, $1400$ of selective search proposals and $3000$ of Randomized Prim's proposals~\citep{randomprime}. %Other methods are not as competitive as those that are reported in the table.

Figure~\ref{fig:iou} {\bf{left}} and {\bf{middle}} show the curves related to recall over \texttt{IoU} with 100 and 1000 proposals. Again, \methodname~obtain good results. The hand crafted segmentation based methods like selective search and MCG have good recall rate at higher \texttt{IoU} values. Instead \methodname~perform better in the range of \texttt{IoU} = [0.6, 0.8] which is desirable in practice and playing an important role in the object detectors performance \citep{Rodrigo14}.

Figure~\ref{fig:iou} ({\bf{right}}) shows average recall(AR) versus number of proposals for different methods. For a specific number of proposals, AR measures the proposal quality across \texttt{IoU} of [0.5, 1]. \citet{Rodrigo14} show that AR correlates well with detection performance. Using this criteria, \methodname~are on par or better than other methods with 700 or fewer boxes but with more boxes, selective search and Edgeboxes perform better.

\paragraph{\textbf{Run-time}}
%\subsubsection{Run-time}
The run-time tests for our proposed method and the others are also available in Table~\ref{table:cmpr}. Since our approach is using the CNN features which are used by state-of-the-art object detectors like RCNN \citep{RCNN} and SppNet \citep{sppnet}, it does not need any extra cues and features and we can consider just the running time of
our algorithm without CNN extraction time \footnote{If CNN features have to be (re)computed, that would add 0.15 sec. extra computation time on our GPU.}. \methodname~takes 0.75 second on CPU and 0.4 second to generate object proposals on a GeForce GTX 750 Ti GPU, which is slightly slower than Edgeboxes. The fastest method is RPN-ZF, a convolutional network based on \citet{zeiler2014visualizing} network architecture, tuned for generating object proposals. Note that for RPN-ZF, the running-time on GPU is reported while the others are reported on CPU. The remaining methods are segmentation based and take considerably more time. 

\begin{table*}[t]
\centering
\scalebox{1}
{
\begin{tabular}{l|*{6}{c}}
			& AUC & N@25\% & N@50\% & N@75\% & Recall  & Time \\
\hline
\hline
BING \citep{BING} 			& .19 & 292 & -   & -    & 29\% & .2s \\
Objectness \citep{objectness}		& .26 & 27  & - & -  & 39\% & 3s\\
Rand. Prim's \citep{randomprime}            & .30 & 42  & 349 & 3023 & 71\% & 1s  \\
Selective Search \citep{selectivesearch}        & .34 & 28  & 199 & 1434 & 79\% & 10s \\
Edge boxes 70 \citep{edgebox}		& .42 & 12  & 108 & 800  & \textbf{84\%} & .3s\\
MCG \citep{MCG}    			& .42 & 9   & 81  & 1363 & 78\% & 30s \\
RPN-ZF \citep{ren2015faster}		& .42 & 13 & 83 & 922 & 78\% & \textbf{.1s}$^*$ \\
\methodname70 		& \textbf{.48} & \textbf{5}   & \textbf{53}  & \textbf{540}  & 83\% & .75s \\
\end{tabular}
}
\caption{Our method compared to other methods for \texttt{IoU} threshold of 0.7. AUC is the area under recall vs. \texttt{IoU} curve for 1000 proposals. N@25\%, N@50\%, N@75\% are the number of proposals needed to achieve a recall of 25\%, 50\% and 75\% respectively. For reporting Recall, at most 2000 boxes are used. The run-times for the other methods are obtained from~\citet{Rodrigo14}. $^*$In contrast to the other methods, for RPM-ZF the run-time is evaluated on GPU.}
\label{table:cmpr}
\end{table*}

%------------------------------------------------------------------------
\iffalse
\subsection{Generality of the approach}
\label{subsec:generality}
As our approach is based on learning, we verify that the method can generalize also to classes that have not been used in training.
For doing that we evaluate the performance of our method on Imagenet ILSVRC'13~\citep{imagenet} validation set including 200 classes of objects.

\begin{figure}[t]
\begin{center}
\begin{tabular}{c@{}c@{}c@{}c}
\includegraphics[width=0.5\linewidth]{imagenet_np_50-crop.pdf}
&
\includegraphics[width=0.5\linewidth]{imagenet_np_70-crop.pdf}
\end{tabular}
%\vspace{-0.5cm}
\caption{Recall versus number of proposals on the Imagenet 2013 validation set for ({\bf{left}}) IoU threshold 0.5 and ({\bf{right}})IoU threshold 0.7.}
\label{fig:nprop}
\end{center}
\end{figure}
}
\fi
%------------------------------------------------------------------------

%------------------------------------------------------------------------
\paragraph{\textbf{Object Detection Perfomrance}}
%\subsubsection{Object detection Performance}
%\label{subsubsec:detectionRes}
%------------------------------------------------------------------------
%PASCAL VOC 2007 also used for evaluation of CNN-based object detector \citet{RCNN} performance using our object proposal boxes. %AG:RCNN or SPP?
%This section analyses the object detectors performance using detection proposals from \methodname~and other methods on PASCAL VOC 2007. We consider two state-of-the-art detectors: RCNN \citet{RCNN} and SppNet \citet{sppnet}.
%Recently it has become clear (see \citet{Rodrigo14}) that an object proposal method with high recall at $0.5$ IoU does not automatically lead to a good detector when its proposals are used as detection hypotheses. 
In the previous experiments we evaluate our proposal generator with different metrics and show that it is among the best methods in all of them. However, we believe that the best way to evaluate the usefulness of the generated proposals is a direct evaluation of the detector performance. Indeed, recently it has become clear (see \citet{Rodrigo14}) that an object proposal method with high recall at $0.5$ IoU does not automatically lead to a good detector.

%As we aim for a fast and accurate detector proposal approach, in this test we evaluate only methods that can compute the proposals in less than $1$ second and provide relatively good recall in the previous experiments.

Some state-of-the-art detectors at the moment are: RCNN~\citep{RCNN}, SppNet~\citep{sppnet}, fast-RCNN~\citep{girshick15fastrcnn}. % and faster-RCNN~\citep{ren2015faster}. 
All are based on CNN features and use object proposals for localizing the object of interest.
The first uses the window proposals to crop the corresponding regions of the image, compute the CNN features and obtain a classification score for each region. This approach is slow and takes around $10$ sec on a high-end GPU and more than $50$ sec on the GPU used for our experiments (GeForce GTX 750 Ti).
%In this case we cannot reuse of the CNN features for our method, but still, the evaluation of the detector performance with different window proposals can give interesting cues about the best approach for proposals generation.
%Results on VOC 2007 in terms of mean AP are shown in Table~\ref{table:cmprDet}.
%The other CNN-based method with state-of-the-art results is SPP.
SppNet and fast-RCNN instead compute the CNN features only once, on the entire image. Then, the proposals are used to select the sub-regions of the feature maps from where to pull the features.
This allows this approach to be much faster. With these approaches then, we can also reuse the CNN features needed for the generation of the proposal so that the complete detection pipeline can be executed without any pre-computed component roughly in $1$ second on our GPU. 

Concurrently to our method also Faster-RCNN was recently introduced~\citep{ren2015faster}. It uses a Region Proposal Network (RPN) for generating proposals that shares full-image convolutional features with the detection network. 
% Thus, similarly to our approach, also faster-RCNN can produce nearly cost-free region proposals. The main difference is that our approach works off-the-shelf with already trained networks, while RPN should be trained together with the detection network in a alternating fashion.

%\red{ For the experiments, we have fine-tuned fast-RCNN detector by both of our \methodname~and the selective search and for SppNet testing, we use trained detector by Selective Search proposals.}
%We evaluate the detection performance of SppNet \red{and fast-rcnn} on VOC 2007 \red{for 2000 proposals of} our \methodname70. As we are only interested in the relative quality of the proposals \red{for SppNet} we do not learn a bounding box regression and we report raw detection results.
%For SppNet, we obtain a mean average precision of $52.2$. This is lower than the $54.5$ that SppNet obtained with selective search (and also without bounding box regression). However, this is expected because the method has been trained with selective search proposals, so it is adapted to them. Similar behavior has been reported before for other methods \citet{Rodrigo14}.

We compare the detection performance of our \methodname70 with selective search and RPN proposals. For RPN proposals the detector is trained as in the original paper~\citep{ren2015faster} with an alternating procedure, where detector and localization sub-network update the shared parameters alternatively.
Our method and selective search are instead evaluated using a detector fine-tuned with the corresponding proposals, but without any alternating procedure, \ie the boxes remain the same for the entire training. %, so that detector and proposal generator are matched and the comparison is fair. 
The training is conducted using faster-RCNN code on PASCAL VOC 2007 with $2000$ proposals per image. In Fig.~\ref{fig:det_res} we report the detector mean average precision on the PASCAL VOC 2007 test data for different number of used proposals. 
%The difference between the two approaches is quite significant and it appears mostly in a regime with low number of proposals. For instance, when using 100 proposals selective search obtains a mean average precision of $28.1$, while our proposals already reach $53.2$. Also, our proposals reach almost the top performance with only $300$ bounding boxes, while selective search needs more than $2000$ boxes to reach its best performance. This is an important factor when seeking for maximum speed. We believe that this different behavior is due to the fact that our method is supervised to select good object candidates, whereas selective search is not.

The difference of selective search with CNN-based approaches is quite significant and it appears mostly in a regime with low number of proposals. For instance, when using 50 proposals selective search obtains a mean average precision (mAP) of $28.1$, while RPN and our proposals obtains a mAP already superior to $50$. We believe that this different behavior is due to the fact that our method and RPN are supervised to select good object candidates, whereas selective search is not. 

%RPN obtains $54.7$ and our proposals reach $53.2$. %MP:Amir please update the numbers
%For instance, when using 100 proposals selective search obtains a mean average precision of $28.1$, RPN obtains $54.7$ and our proposals reach $53.2$. 

Comparing our proposals with RPN, we observe a similar trend. \methodname~produces superior results with a reduced amount of proposals ($<100$), while RPN performs better in the range of between $100$ and $700$ proposals. With more than $700$ proposals both methods perform again similarly and better than selective search. %, reaching a detection performance of almost $60$ mAP. 
Finally, with $2000$ proposals per image, selective search, RPN and \methodname~ reach to the detection performance of $59.3$, $59.4$ and $59.8$ respectively.

%RPN and our proposals reach almost the top performance with only $300$ bounding boxes, while selective search needs more than $2000$ boxes to reach its best performance. This is an important factor when seeking for maximum speed. We believe that this different behavior is due to the fact that our method is supervised to select good object candidates, whereas selective search is not. Also, we note that RPN is originally proposed to be integrated with faster-RCNN detector. However, to have a fair comparison with ours and selective search proposals, we have used RPN in combination with fast-RCNN detector. %MP: not clear what you do exactly...

% Thus, similarly to our approach, also faster-RCNN can produce nearly cost-free region proposals. The main difference is that our approach works off-the-shelf with already trained networks, while RPN should be trained together with the detection network in a alternating fashion.
Thus, from these results we can see that RPN and our approach perform very similar. The main difference between the two approaches lies in the way they are trained. Our approach assumes an already pre-trained network, and learns to localize the object of interest by leveraging the convolutional activations generated for detection. RPN instead needs to be trained together with the detector with an alternating approach. In this sense, our approach is more flexible because it can be applied to any CNN based detector without modifying its training procedure.

%Using a pretrained SppNet fine-tuned for selective search, we obtain a mAP of 52.2 with \methodname~which is lower than 54.5 of the selective search. Similar behavior has been reported for other methods and can be explained by the use of different proposals for training and testing~\citep{Rodrigo14}.

%An advantage of our approach is that it can obtain higher recall than other methods for a reduced set of proposals (see Fig.~\ref{fig:iou}~({\bf right})). In this setting, even if the SppNet has been trained with selective search proposals, our method obtains better detection accuracy. More specifically, testing SppNet using only $100$ proposals our method attains an mAP of $46.5$ whereas selective search obtains $45.5$. Furthermore, our approach with a reduced set of proposals can be faster. This is not the case for  selective search, being based on a fixed-cost segmentation procedure.

%MP: shall we keep this? It is true only for selective search. I would remove it...
%An additional advantage of our approach, being based on learning, is that it can focus on specific classes. To illustrate this, we train a special version of \methodname~for cars, where the positive training samples are collected only from car instances.
%In this setting the performance of the car detector improves from $57.6\%$ to $60.4\%$ using SppNet. Thus, in this scenario, our proposals can also be used to improve a detector performance.

\begin{figure}[t]
\centering
\small
\scalebox{0.7}
{
	\includegraphics[width=1\linewidth]{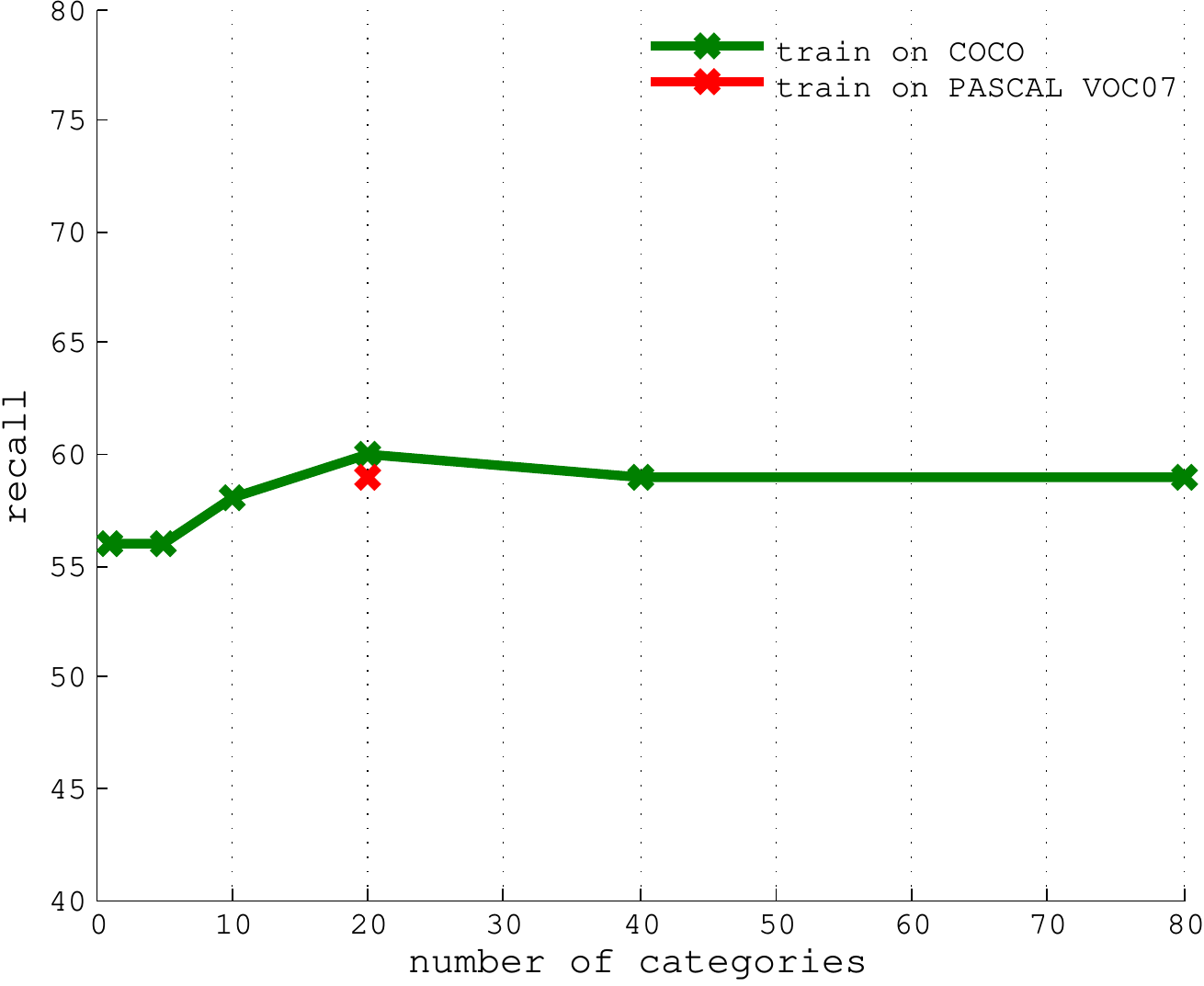}
}
\caption{Generalization of \methodname: we train models with different number of categories and evaluate it on whole eval-set of COCO dataset. We set IoU threshold to $0.5$ and number of proposals to $1000$.}
\label{fig:gener}
\end{figure}

%------------------------------------------------------------------------
\paragraph{\textbf{Generalization to Unseen Categories}}
%\subsubsection{Generalization to unseen categories}
%\label{subsubsec:generalization}
%------------------------------------------------------------------------
We evaluate the generalization capability of our approach on the Microsoft COCO dataset \citep{coco}. The evaluation of the approach has been done by learning either from the 20 classes from VOC07 or COCO or from 1, 5, 20, 40, or 80 categories randomly sampled from COCO. 
As shown in figure~\ref{fig:gener}, when the \methodname~are trained using only 5 classes, the recall at $0.5$ IoU with 1000 proposals is slightly reduced ($56\%$).
With more classes, either using VOC07 or COCO, recall remains stable around $59\%$ - $60\%$. 
This shows that the method can generalize well over all classes. We believe this is due to the simplicity of the classifier (average pooling on CNN features) that avoids over-fitting to specific classes. 
Note that in this case our recall is slightly lower than the Selective Search with 1000 proposals ($63\%$).
This is probably due to the presence of very small objects in the COCO dataset, that are missed by our method as it was not tuned for this setting.
These results on COCO demonstrate that our proposed method is capable to generalize learnt objectness beyond the training categories.

\begin{figure*}[h]
\centering
\small
\scalebox{0.9}

\begin{tabular}{c@{}|c@{}|c@{}|c@{}}
DeepProposals-70 & EdgeBoxes-70 & DeepProposal-70 & EdgeBoxes-70

\\

\includegraphics[width=4cm, height=3cm]{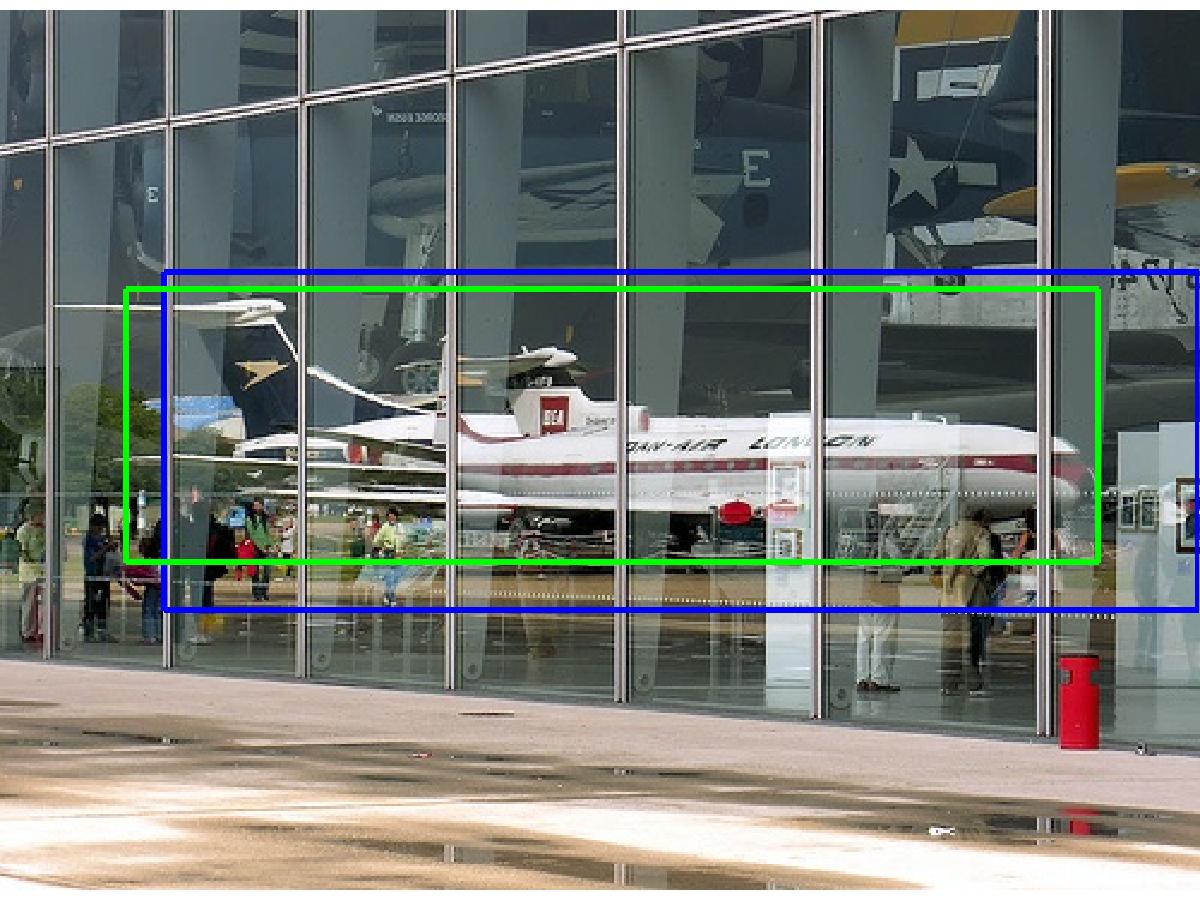}
&
\includegraphics[width=4cm, height=3cm]{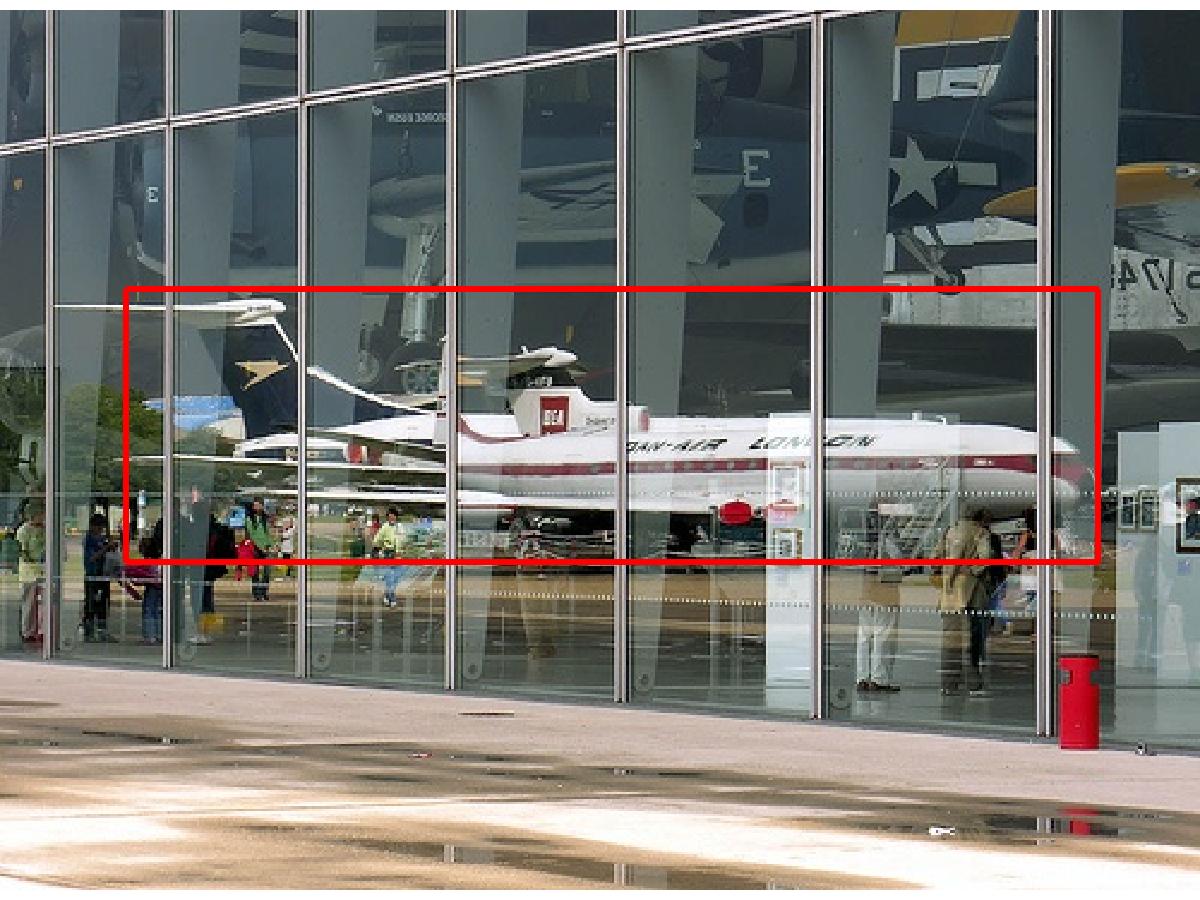}
&
\includegraphics[width=4cm, height=3cm]{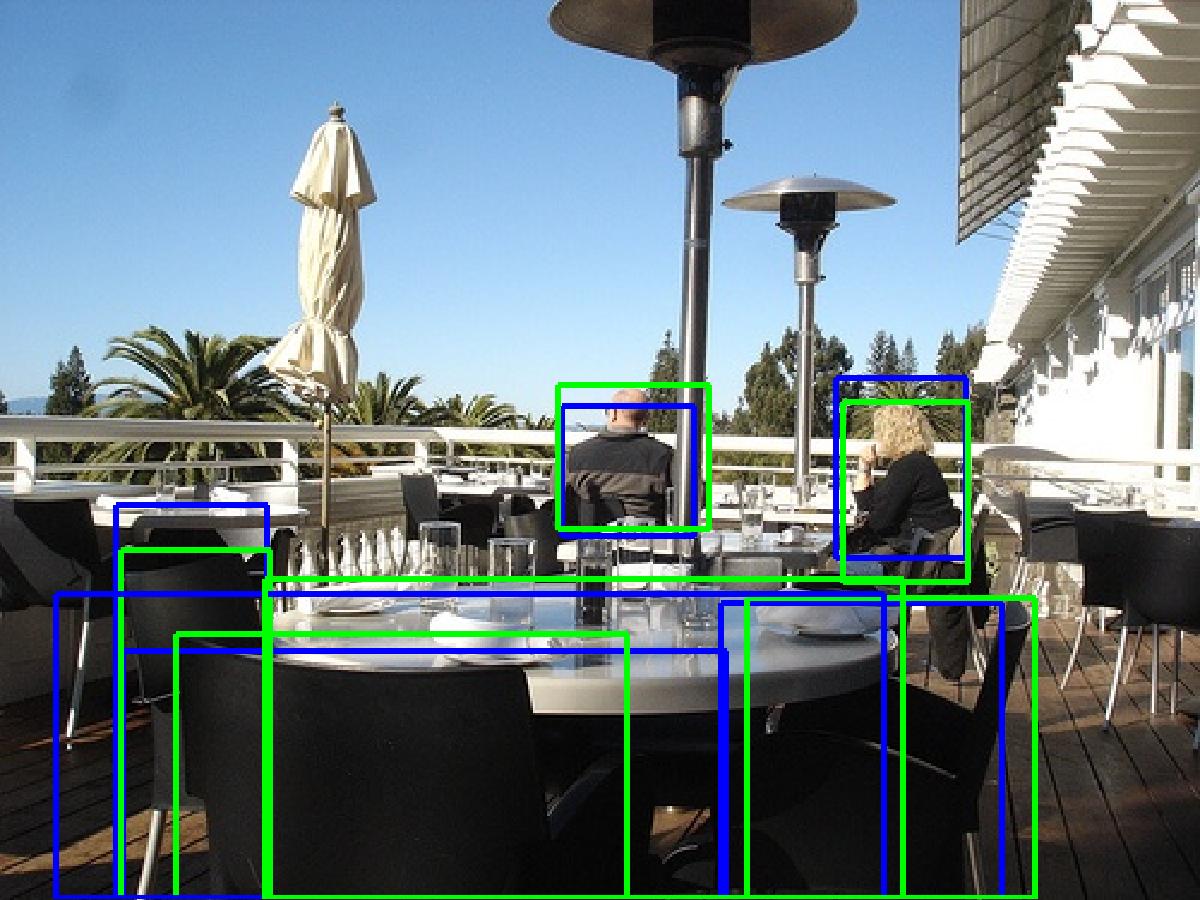} 
&
\includegraphics[width=4cm, height=3cm]{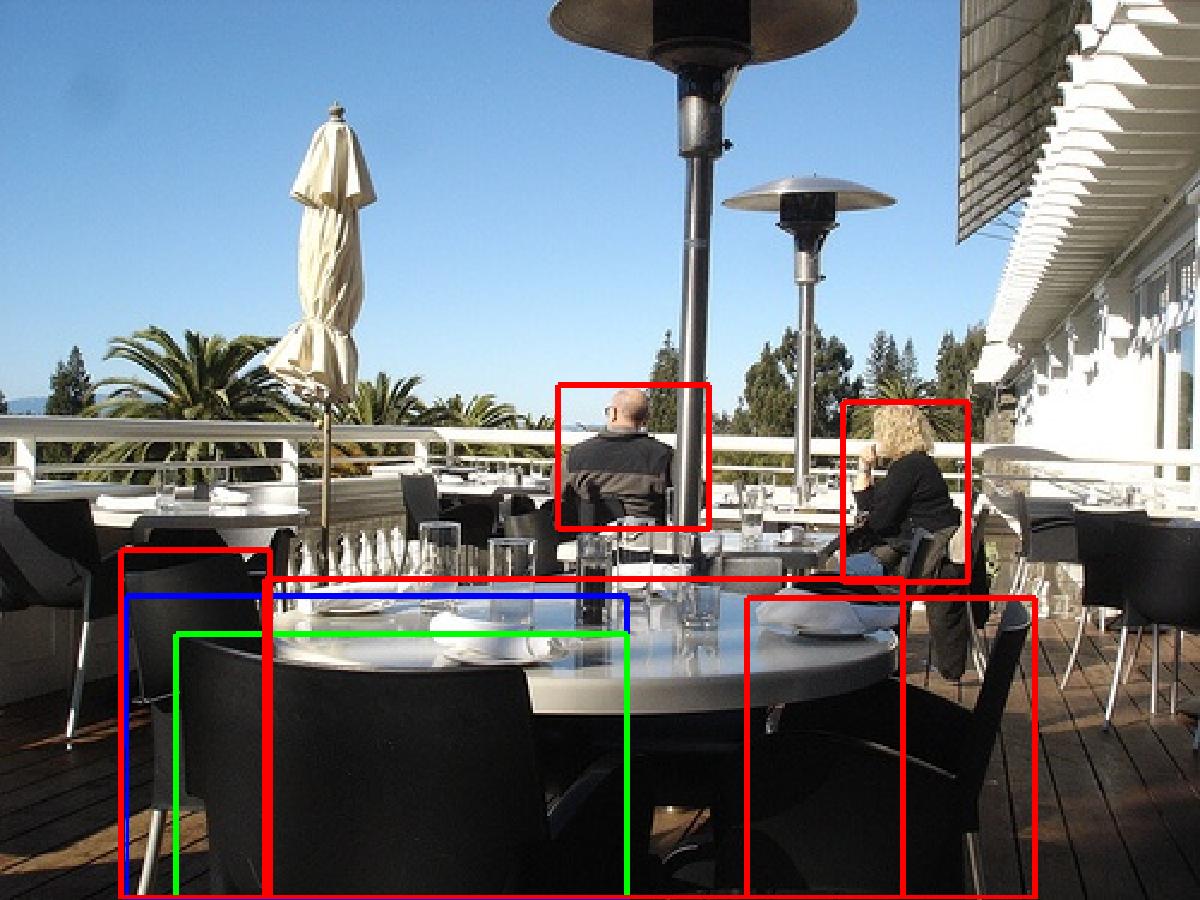}

\\

\includegraphics[width=4cm, height=3cm]{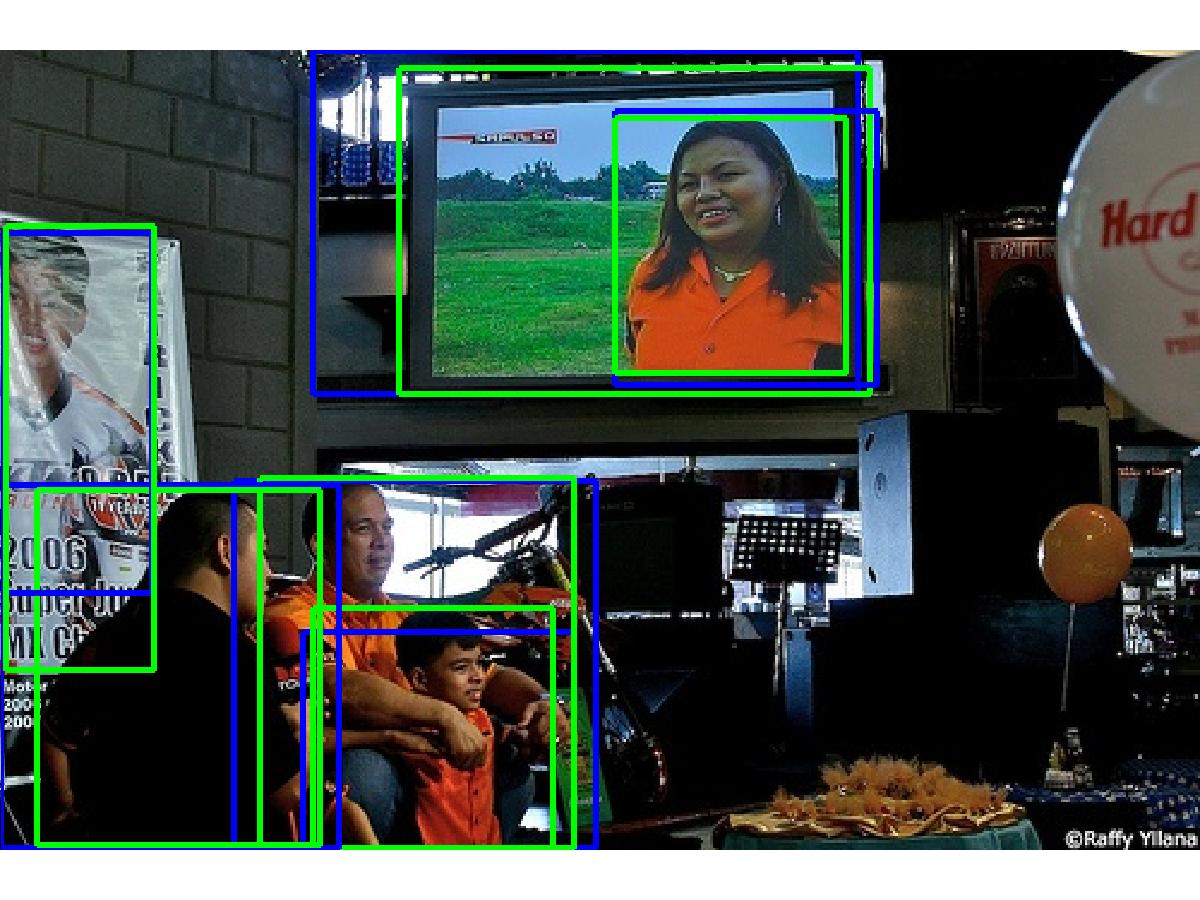}
&
\includegraphics[width=4cm, height=3cm]{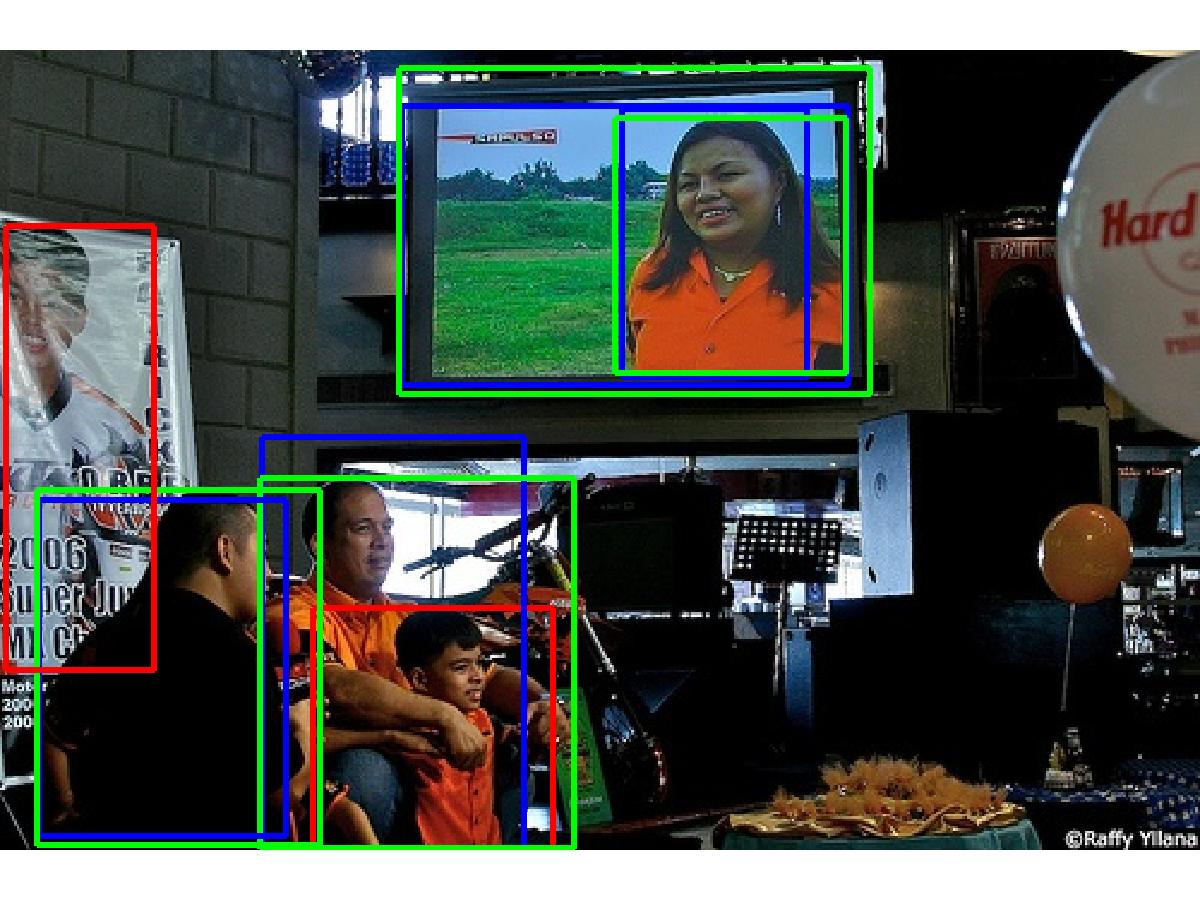}
&
\includegraphics[width=4cm, height=3cm]{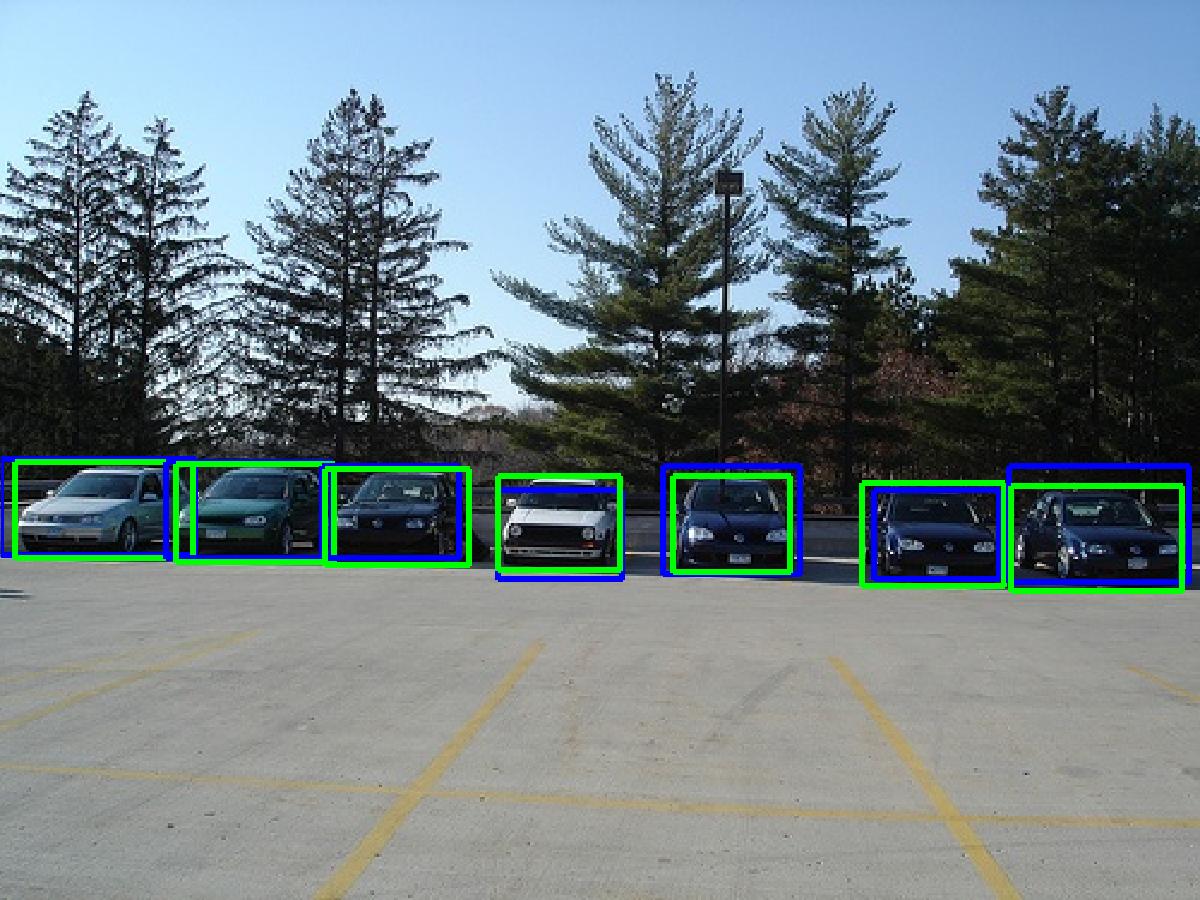}
&
\includegraphics[width=4cm, height=3cm]{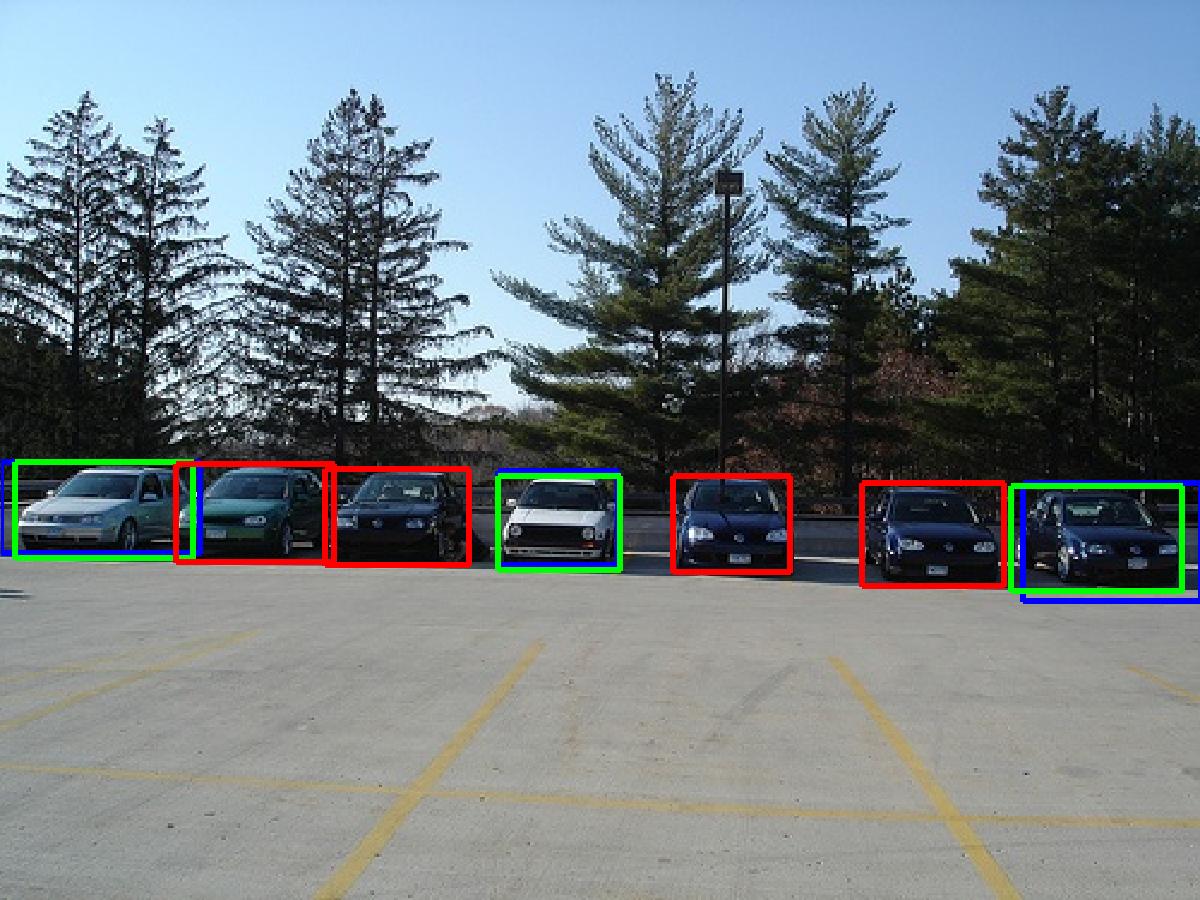}

\\

\includegraphics[width=4cm, height=3cm]{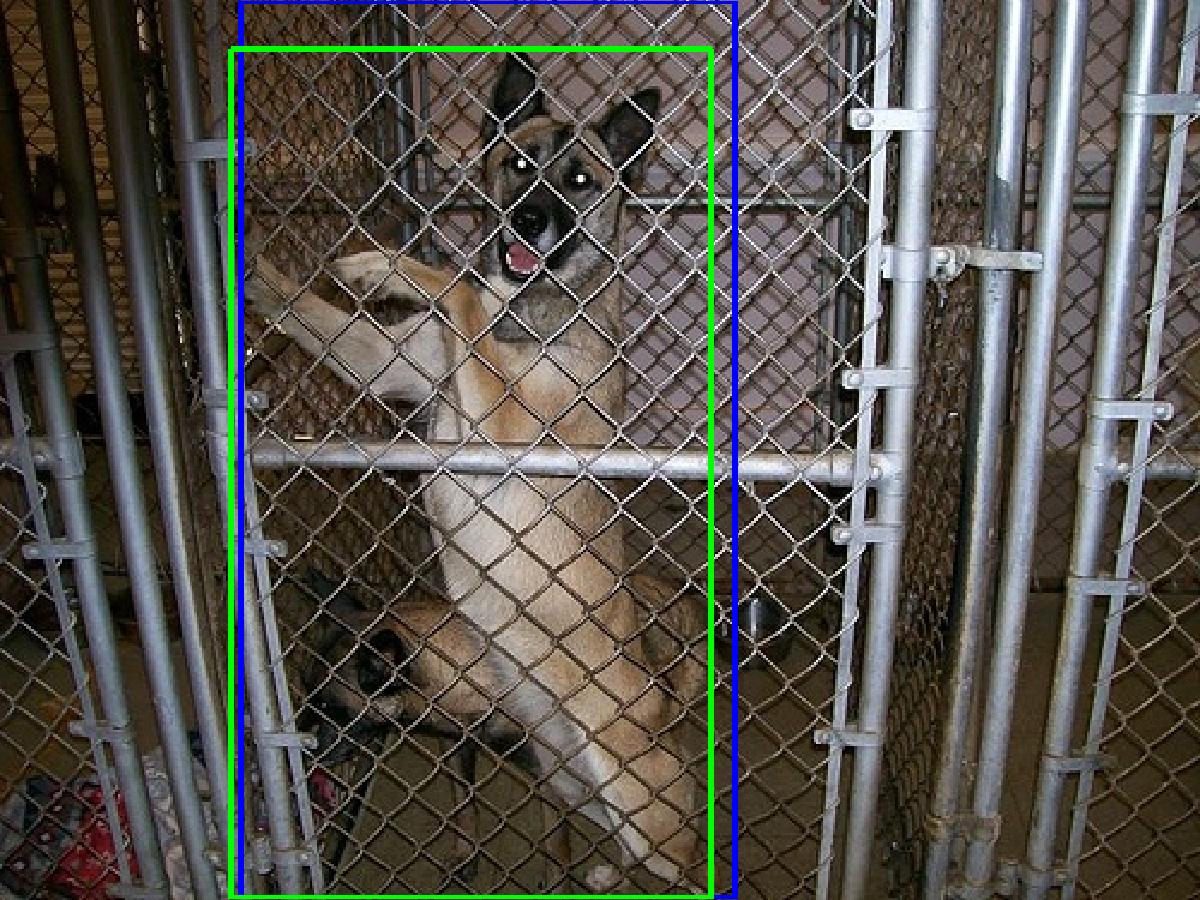}
&
\includegraphics[width=4cm, height=3cm]{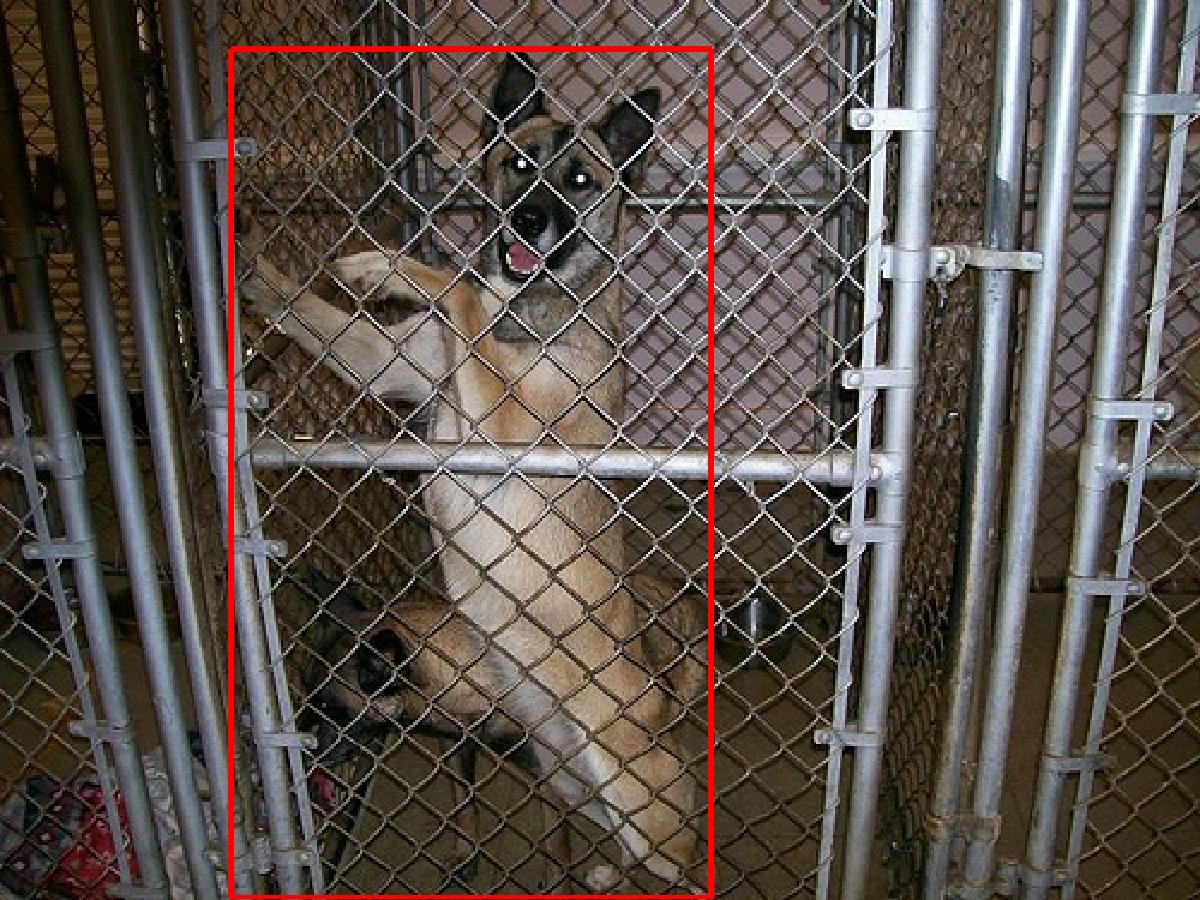}
&
\includegraphics[width=4cm, height=3cm]{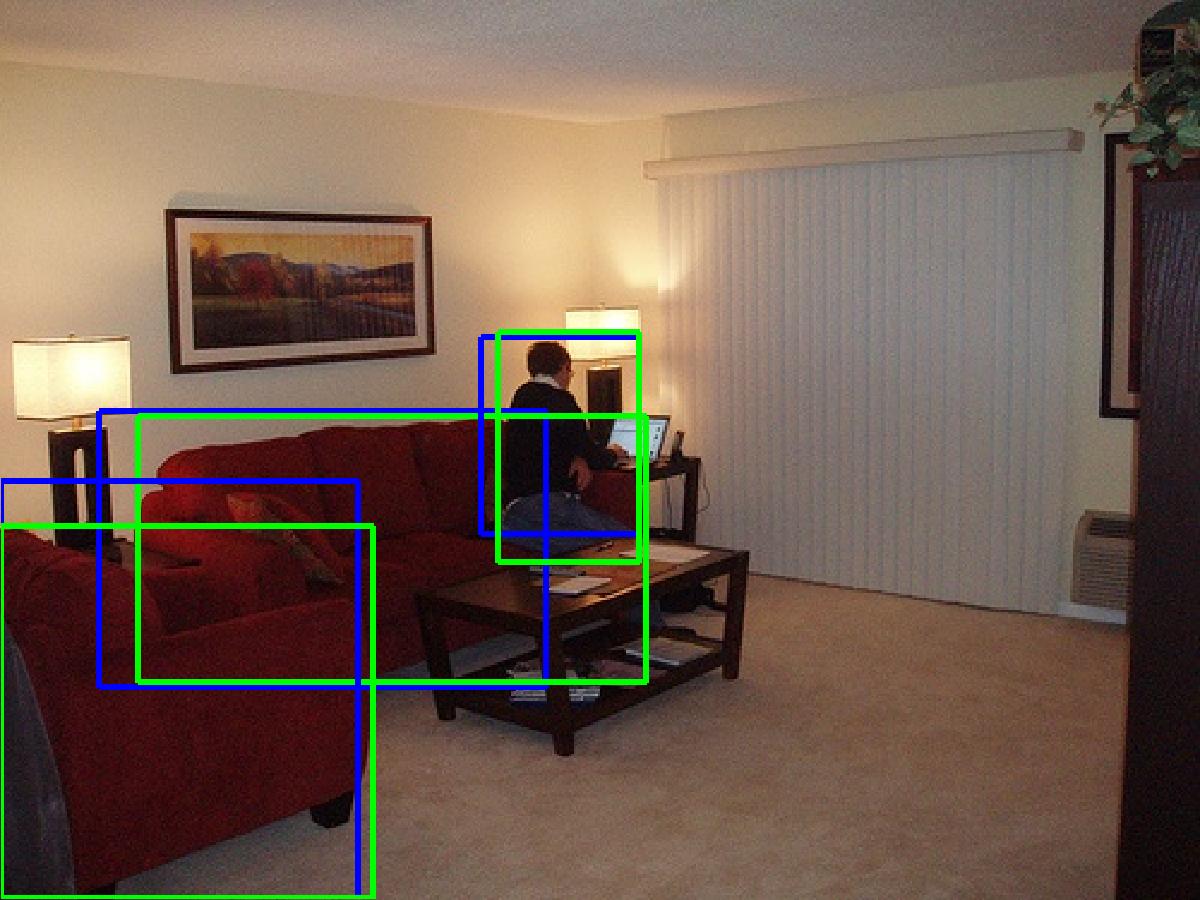} 
&
\includegraphics[width=4cm, height=3cm]{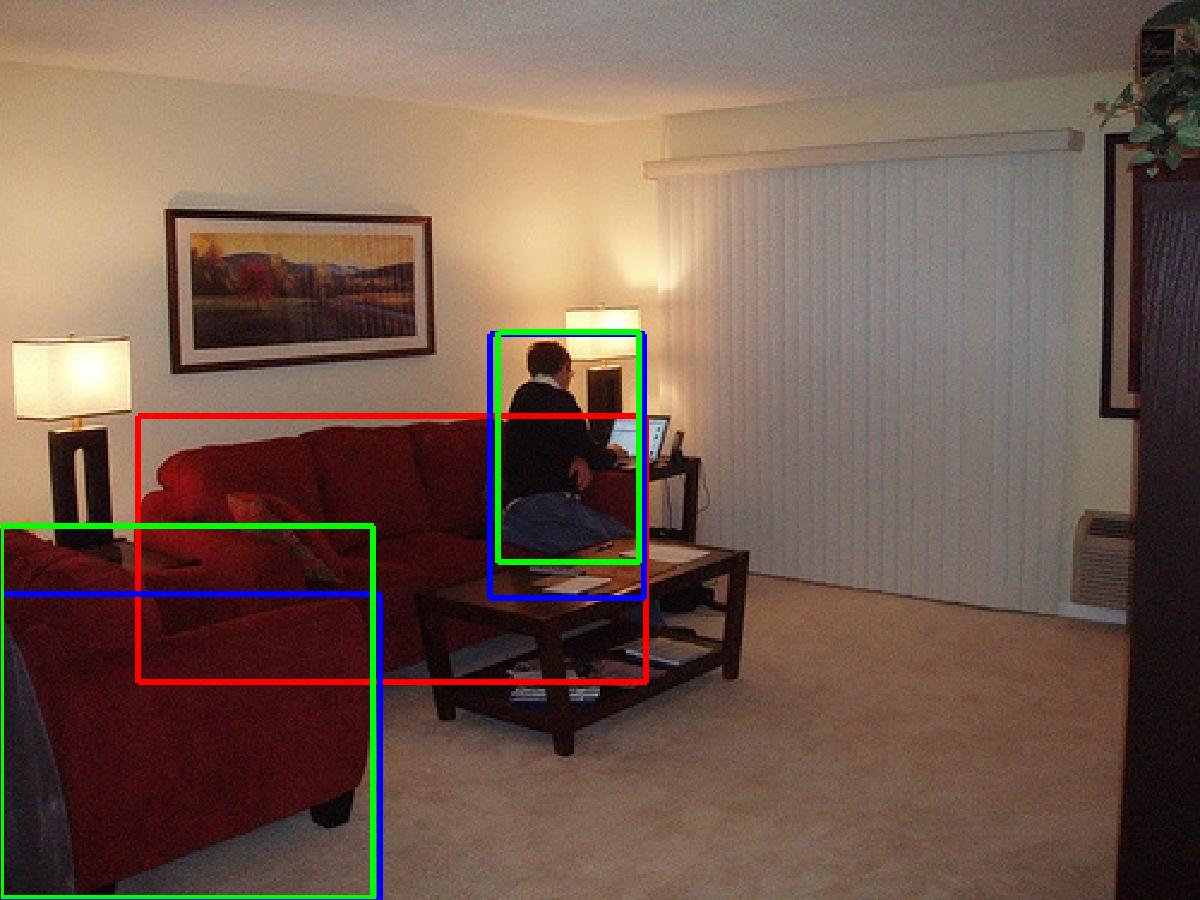}

\\

%\includegraphics[width=0.24\linewidth]{images/samples/000179_db.jpg}
%&
%\includegraphics[width=0.24\linewidth]{images/samples/000179_eb.jpg}
%&
%\includegraphics[width=0.24\linewidth]{images/samples/000201_db.jpg} 
%&
%\includegraphics[width=0.24\linewidth]{images/samples/000201_eb.jpg}
%
%\\
%
\includegraphics[width=4cm, height=3cm]{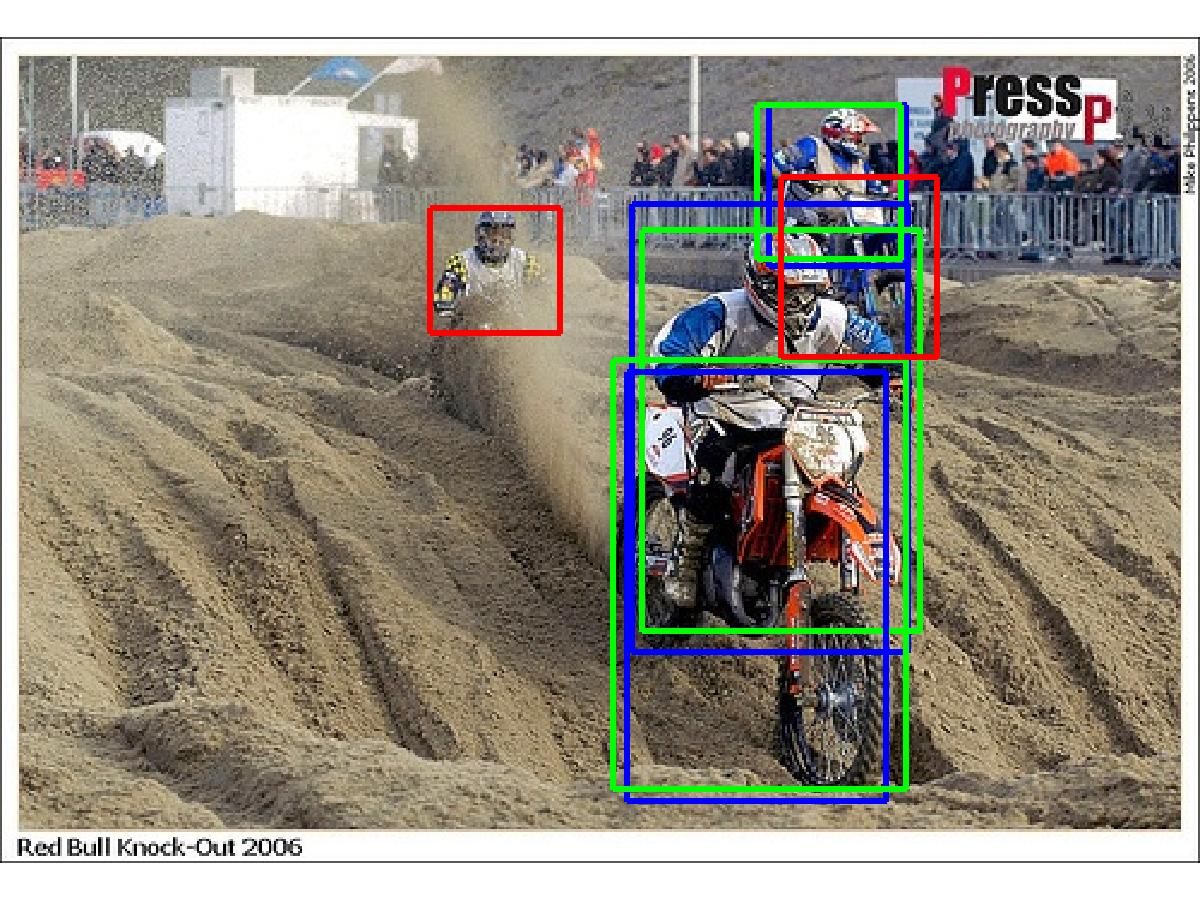}
&
\includegraphics[width=4cm, height=3cm]{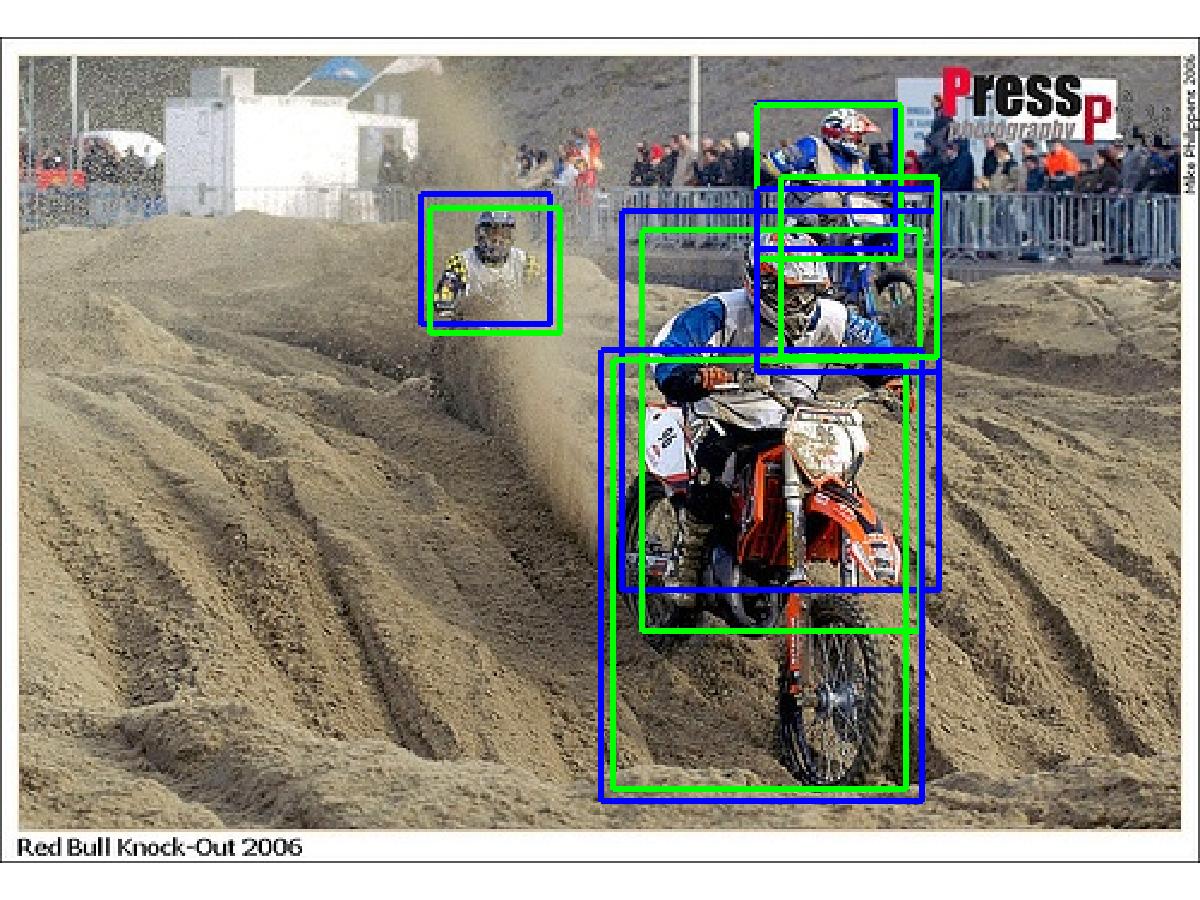}
&
\includegraphics[width=4cm, height=3cm]{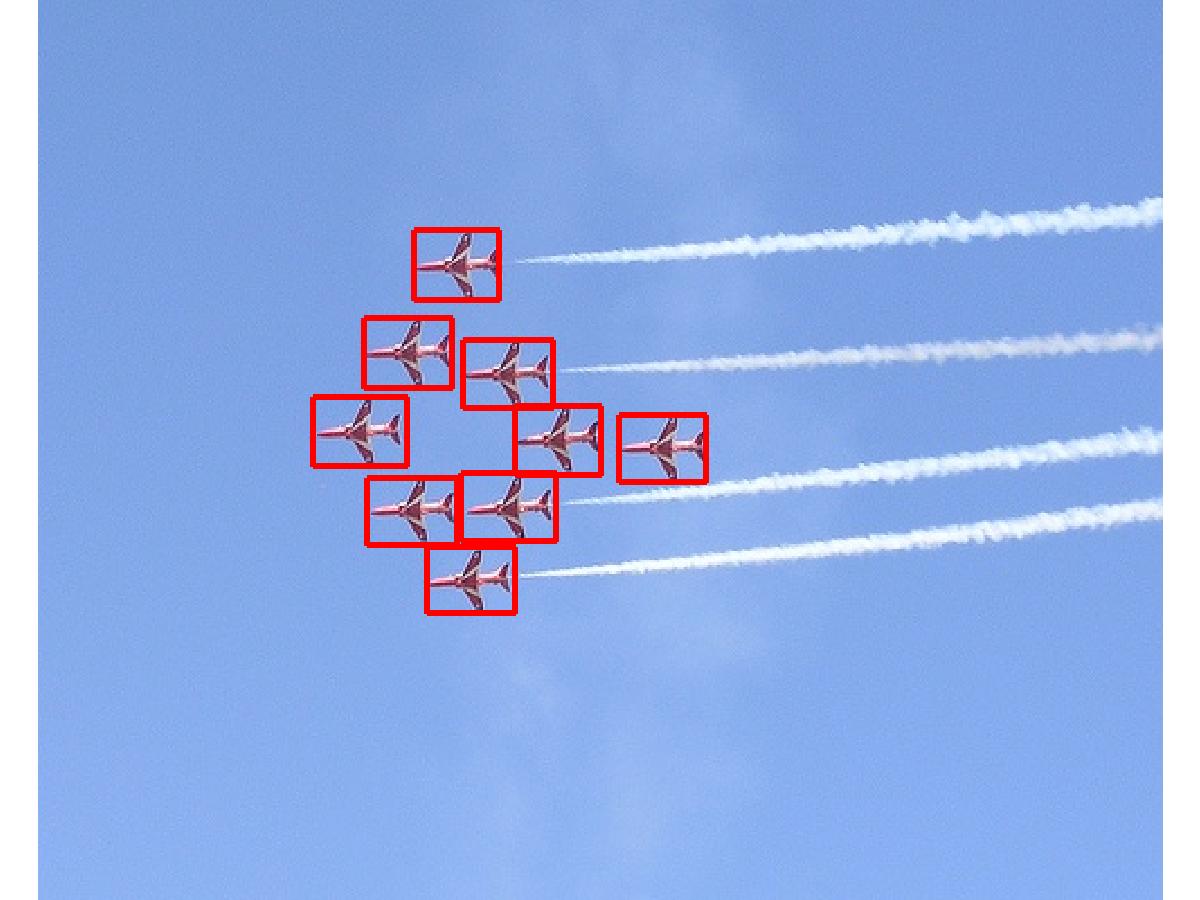} 
&
\includegraphics[width=4cm, height=3cm]{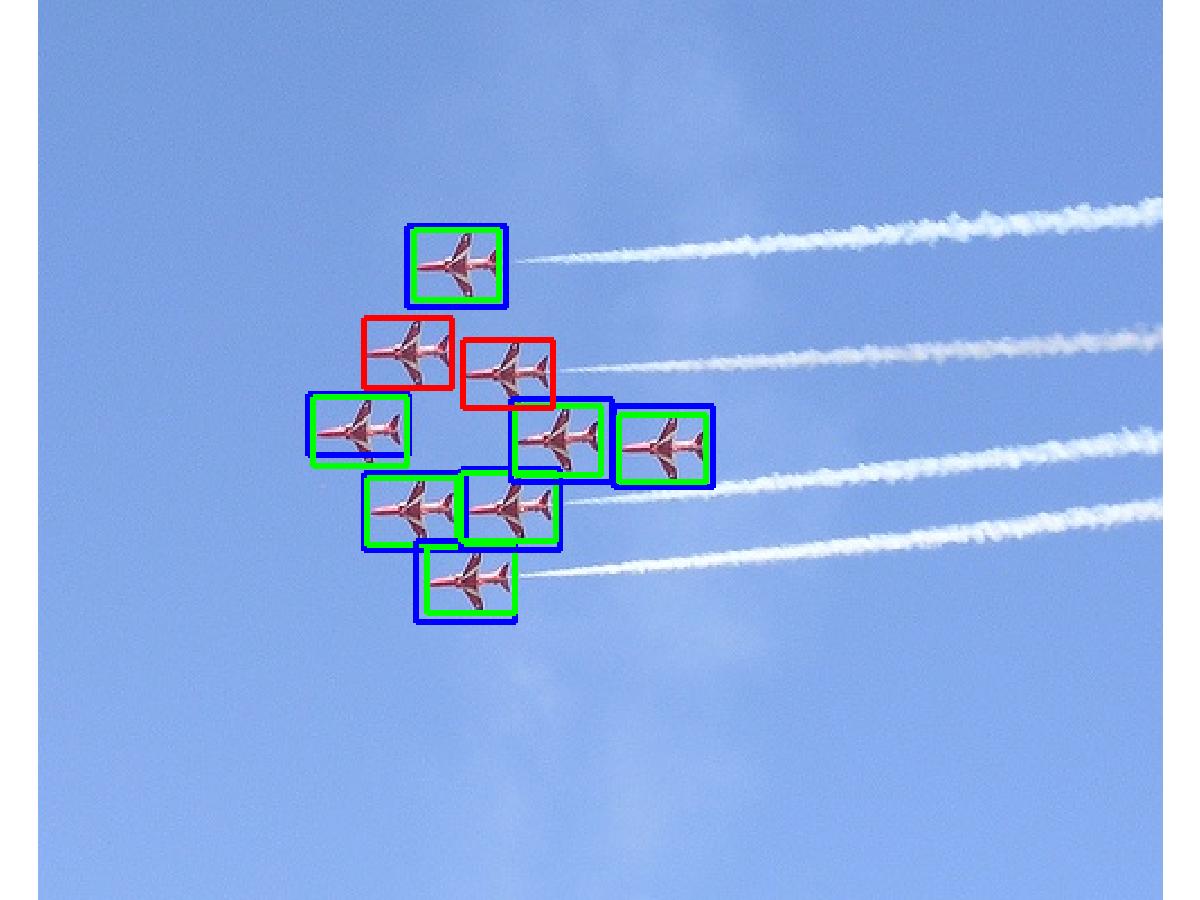}

\end{tabular}
\caption{Qualitative examples of our object proposals (1st and 3rd column) versus Edge boxes proposals (2nd and 4th column). For first three rows our method performs better while in the last row Edge boxes is better. An object is correctly localized if its \texttt{IoU} with the ground-truth bounding box is more than $0.7$. We use {\bf{1000}} proposals for each method. Blue boxes are the closest proposal to each ground truth bounding box. Red and green boxes are ground-truth boxes where green indicates a localized object while red indicates a missed object.}
\label{fig:qualitative}
\end{figure*}

%------------------------------------------------------------------------
\paragraph{\textbf{Qualitative Results}}
%\subsubsection{Qualitative Results}
%\label{subsubsec:qualitative}
%------------------------------------------------------------------------
Figure \ref{fig:qualitative} shows some qualitative results of \methodname~and another state of the art method, Edge boxes. In general,
when the image contains high-level concepts cluttered with many edges (e.g. Figure~\ref{fig:qualitative} rows 1 and 3, first column) our method gives better results. However, for small objects with clear boundaries edge boxes performs better since it is completely based on contours and can easily detect smaller objects.

%------------------------------------------------------------------------
\subsection{Action Proposals}
\label{subsec:actionp}
%------------------------------------------------------------------------
We evaluate our action proposals on two different datasets namely UCF-Sports \citep{rodriguez2008action} and UCF101 \citep{soomro2012ucf101}. UCF-Sports contains 10 action categories and consists of 150 video samples, extracted from sport broadcasts. The actions in this dataset are temporally trimmed. For this dataset we use the train and test split proposed in \citet{lan2011discriminative}. UCF101 is collected from YouTube and has 101 action categories where 24 of the annotated classes (corresponding to $3,204$ videos) are used in literature for action localization. %Only 74.6\% of videos are temporally trimmed in this dataset. Therefore, we apply the Viterbi algorithm as explained in~\ref{subsec:actionp} for temporal window sizes of $[20, 40, 60, 80, 100, 150, 170]$ frames with step size of $30$ frames.
%We train our models on training set of split-3 of UCF101 ($2,303$ videos) and evaluate our method on test set of split-3 ($901$ videos).
In this dataset, for evaluation we report the average recall of 3 splits. %as the performance. To report the performance of other methods, exactly the same protocol as us is used for both datasets.
Finally, for both datasets, we select the first top $100$ boxes in each frame and find the $N$ best paths over time for each video.

%------------------------------------------------------------------------
\paragraph{\textbf{Comparison with the state-of-the-art}}
%\subsubsection{Comparison with state-of-the-art}
%\label{subsubsec:comparison_action}
%------------------------------------------------------------------------
In Table \ref{table:cmpr_a} we compare our proposal generation method against state-of-the-art methods in the presented datasets. As shown, our method is competitive or improves over all other methods with fewer proposals. In the UCF-Sports dataset, \methodname~have higher recall compared to the recently published APT proposal generator \citep{van2015apt} with almost $70$x fewer proposals. Notice that the method proposed by \citet{brox2010object} is designed for motion segmentation and we use it here to emphasize the difficulty of generating good video proposals.
%In UCF101 dataset we are on par with APT while using $24$ times less proposals and outperform it when we use $500$ video proposals in each video ($4.8$x less than APT).
In the UCF101 dataset we see the same trend, we outperform APT while using $67$x fewer proposals.

\paragraph{\textbf{Run-time}}
Computationally, given the optical flow images, our method needs $1.2$ seconds per-frame to generate object proposals and on average $1.3$ seconds for linking all the windows. Most of the other methods %MP: it is a bit too general; are there other fast methods?
are generally order of magnitude more expensive mainly because of performing super-pixel segmentation and grouping.

\begin{table}[t]
\centering
\begin{tabular}{l|*{2}{c}}
			& Recall & \#proposals \\
\hline
\textbf{UCF-Sports} &  &  \\
\citet{brox2010object} & 17.02 & \textbf{4} \\
\citet{selectivesearch_v} & 78.72 & 1642 \\on average 
\citet{randprime_v} & 68.09 & 3000 \\
\citet{gkioxari2015finding} & 87.23 & 100 \\
APT~\citep{van2015apt} & 89.36 & 1449 \\
\methodname & \textbf{95.7} & 20\\
\hline
\textbf{UCF101} &  &  \\
%APT~\citet{van2015apt} & 40 & 2400 \\
%DeepProposal &  40 & \textbf{100} \\
%DeepProposal &  \textbf{43} & 500 \\
APT~\citep{van2015apt} & 37.0 & 2304 \\
\methodname &  \textbf{38.6} & \textbf{34} \\
\end{tabular}
\caption{Our action proposals generator compared to other methods at \texttt{IoU} threshold of $0.5$. The number of proposals is averaged over all test videos. All the reported numbers on UCF-sports except ours are obtained from~\citet{van2015apt}. For UCF101, like ours, we report the APT performance for split3. }
\label{table:cmpr_a}
\end{table}

\begin{figure*}
%\centering
\begin{center}
%\scalebox{0.6}
{
\includegraphics[width=1\linewidth]{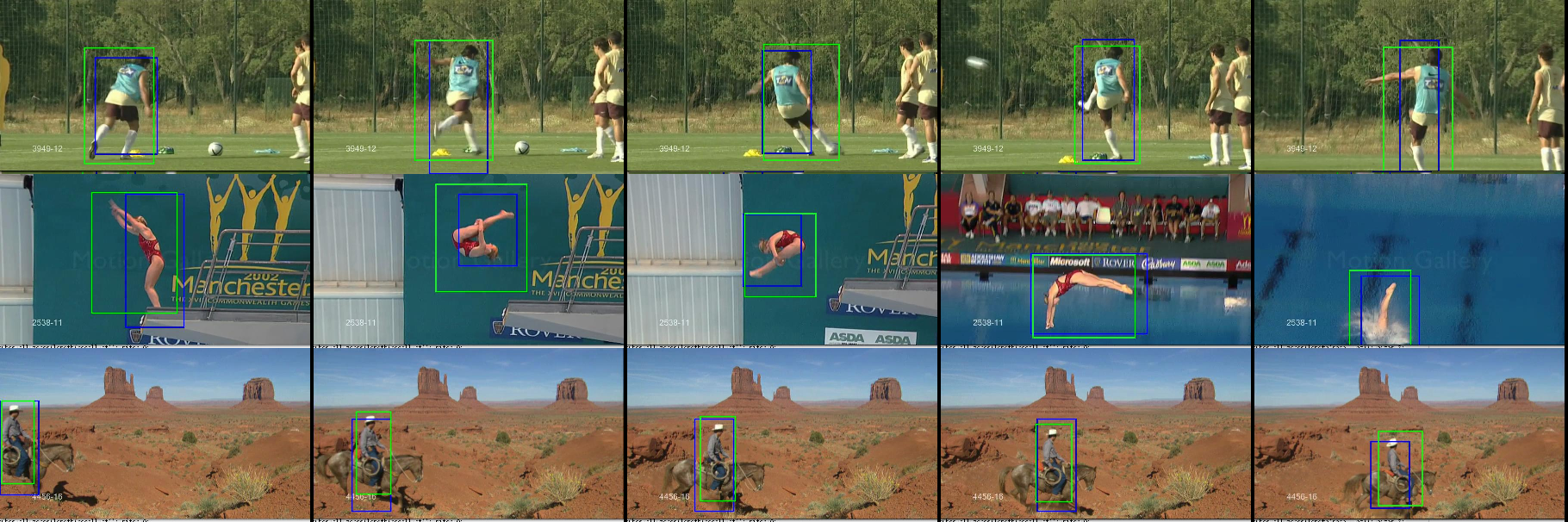}
}
\end{center}
\caption{Qualitative examples of our action proposals. In each row, we select 5 frames of a video for visualization purpose. Blue boxes are the closest proposal to each ground truth box (shown in green). }
\label{fig:qualitative_a}
\end{figure*}

%------------------------------------------------------------------------
\paragraph{\textbf{Qualitative Results}}
%\subsubsection{Qualitative Results}
%\label{subsubsec:qualitative_action}
%------------------------------------------------------------------------
In figure~\ref{fig:qualitative_a} we provide examples of our action proposals extracted from some videos of UCF-sports dataset. For each video we show 5 cross sections of a tube. These sections are equally distributed in the video.
%------------------------------------------------------------------------
\section{Conclusion}
\label{sec:conclusion}
%------------------------------------------------------------------------
\methodname is a new method to produce fast proposals for object detection and action localization. In this paper, we have presented how \methodname~produces proposals at low computational cost through the use of an efficient coarse to fine cascade on multiple layers of a detection network, reusing the features already computed for detection. We have accurately evaluated the method in the most recent benchmarks and against previous approaches and we have shown that in most of the cases it is comparable or better than state-of-the-art approaches in terms of both accuracy and computation. 
The source code of \methodname~is available online\footnote{https://github.com/aghodrati/deepproposal}.

%------------------------------------------------------------------------
% Acknowledgements
%------------------------------------------------------------------------
\begin{acknowledgements}
This work was supported by DBOF PhD scholarship, KU Leuven CAMETRON project and FWO project ``Monitoring of Abnormal Activity with Camera Systems''.
\end{acknowledgements}

%------------------------------------------------------------------------
% Bibliography
\bibliographystyle{spbasic}      % basic style, author-year citations
\bibliography{bibliography}

\end{document}
% end of file template.tex